\PassOptionsToPackage{dvipsnames}{xcolor}
\documentclass[]{bytedance_seed}

\usepackage{minitoc}
\usepackage{cleveref} 
\usepackage{subcaption}
\usepackage{booktabs}

\usepackage{natbib}
\usepackage{latexsym}

\usepackage[utf8]{inputenc}    
\usepackage[T1]{fontenc}       
\usepackage{microtype}         

\usepackage{amssymb}           
\usepackage{amsmath}           
\usepackage{amsfonts}          
\usepackage{amsthm}            

\usepackage{booktabs}          
\usepackage{multirow}          
\usepackage{makecell}          
\usepackage{colortbl}          
\usepackage[dvipsnames]{xcolor}           

\usepackage{graphicx}          
\usepackage{tikz}              
\usepackage{pgfplots}          
\usepgfplotslibrary{groupplots} 
\usepackage[edges]{forest}     
\usepackage{subcaption} 

\usepackage{algorithm}         
\usepackage{algorithmic}       

\usepackage{wrapfig}           
\usepackage{caption}           
\usepackage{adjustbox}         

\usepackage{paralist}          

\usepackage{pifont}            

\usepackage{hyperref}          
\usepackage{url}               

\usepackage[toc,page,header]{appendix} 

\usepackage{xspace}            
\usepackage{nicefrac}          

\hypersetup{
    colorlinks,
    linkcolor={blue!80!black},
    citecolor={blue!80!black},
    urlcolor={blue!80!black}
}
\tikzset{
    root/.style =             {align=center, text width=1cm, rounded corners=3pt, line width=0.3mm, fill=gray!10, draw=gray!80, font=\small},
    demographic/.style =         {align=center, text width=1.8cm, rounded corners=3pt, line width=0.3mm, fill=blue!10, draw=blue!80, font=\footnotesize},
    demographic_work/.style =    {align=center, text width=10cm, rounded corners=3pt, line width=0.3mm, fill=blue!10, draw=blue!0, font=\footnotesize},
    character/.style =         {align=center, text width=1.8cm, rounded corners=3pt, line width=0.3mm, fill=red!10, draw=red!80, font=\footnotesize},
    character_work/.style =    {align=center, text width=10cm, rounded corners=3pt, line width=0.3mm, fill=red!10, draw=red!0, font=\footnotesize},
    personalization/.style =           {align=center, text width=1.8cm, rounded corners=3pt, line width=0.3mm, fill=cyan!10, draw=cyan!80, font=\footnotesize},
    personalization_work/.style =      {align=center, text width=10cm, rounded corners=3pt, line width=0.3mm, fill=cyan!10, draw=cyan!0, font=\footnotesize},
    risk/.style =         {align=center, text width=1.8cm, rounded corners=3pt, line width=0.3mm, fill=orange!10, draw=orange!80, font=\footnotesize},
    risk_work/.style =    {align=center, text width=10cm, rounded corners=3pt, line width=0.3mm, fill=orange!10, draw=orange!0, font=\footnotesize},
}

\newcommand{\method}{\textsc{Enigmata}\xspace}

\usepackage{CJK}
\usepackage{minitoc}

\usepackage{cleveref}

\usepackage{array}

\title{\method: Scaling Logical Reasoning in Large Language Models with Synthetic Verifiable Puzzles}

\affiliation[1]{ByteDance Seed}
\affiliation[2]{Fudan University\quad $^3$Institute for AI Industry Research (AIR), Tsinghua University}
\affiliation[4]{Nanjing University\quad $^5$Shanghai Jiao Tong University}
\affiliation[6]{SIA-Lab of Tsinghua AIR and ByteDance Seed}

\contribution{Full author list in \nameref{sec:contributions}.}

\abstract{
Large Language Models (LLMs), such as OpenAI's o1 and DeepSeek's R1, excel at advanced reasoning tasks like math and coding via Reinforcement Learning with Verifiable Rewards (RLVR), but still struggle with puzzles solvable by humans without domain knowledge.
We introduce \method, the first comprehensive suite tailored for improving LLMs with puzzle reasoning skills. 
It includes 36 tasks across seven categories, each with 1) a generator that produces unlimited examples with controllable difficulty and 2) a rule-based verifier for automatic evaluation.
This generator-verifier design supports scalable, multi-task RL training, fine-grained analysis, and seamless RLVR integration.
We further propose \method-Eval, a rigorous benchmark, and develop optimized multi-task RLVR strategies.
Our trained model, Qwen2.5-32B-\method, consistently surpasses o3-mini-high and o1 on the puzzle reasoning benchmarks like \method-Eval, ARC-AGI (32.8\%), and ARC-AGI 2 (0.6\%).
It also generalizes well to out-of-domain puzzle benchmarks and mathematical reasoning, with little multi-tasking trade-off.
When trained on larger models like Seed1.5-Thinking (20B activated parameters and 200B total parameters), puzzle data from \method further boosts SoTA performance on advanced math and STEM reasoning tasks such as AIME (2024-2025), BeyondAIME and GPQA (Diamond), showing nice generalization benefits of \method.
This work offers a unified, controllable framework for advancing logical reasoning in LLMs. 

}

\correspondence{Jiangjie Chen at \email{jiangjiec@bytedance.com}}

\checkdata[Project Page]{\url{https://seed-enigmata.github.io/}}

\begin{document}

\maketitle

\section{Introduction}

\begin{figure}[h]
    \centering
    \includegraphics[width=1\linewidth]
    {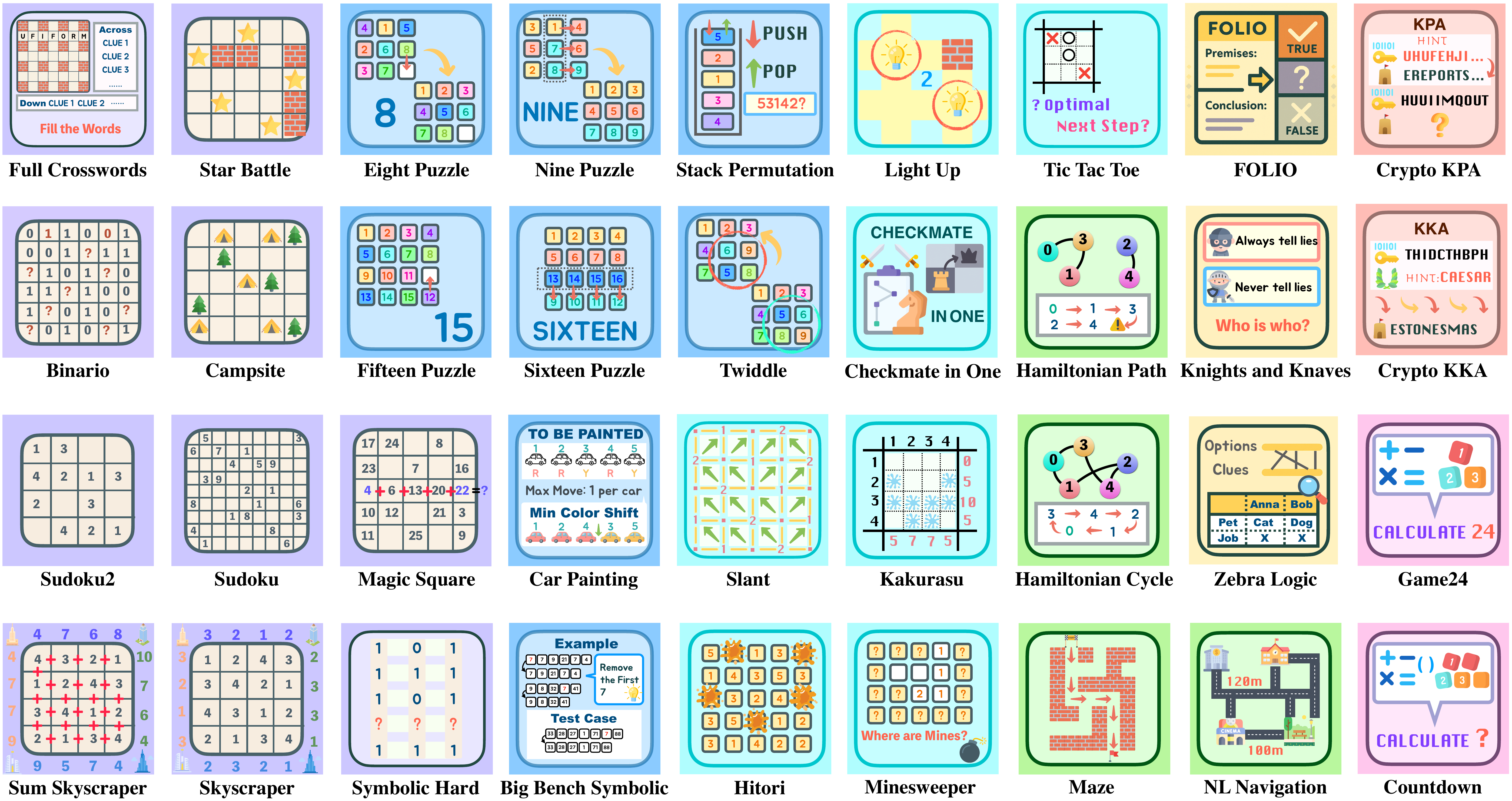}
    \caption{Overview of the \method dataset: 36 puzzle tasks across seven categories, designed to enhance and evaluate diverse reasoning capabilities in large language models.}
    \label{fig:data}
\end{figure}
\vspace{-3pt}

Large Reasoning Models (LRMs) such as o1 and R1, trained from Large Language Models (LLMs) with Reinforcement Learning (RL), have demonstrated excellent performance in complex reasoning tasks~\cite{o1,guo2025deepseek,gemini,claude,seed2025seed}, such as mathematics, STEM, and coding. 
The success of LRMs highlights the effectiveness of the Reinforcement Learning with Verifiable Rewards (RLVR) paradigm, which stresses the great importance of
obtaining high-quality verifiable prompts~\cite{xu2025kodcode, albalak2025big, he2025deepmath}.
However, existing LRMs still struggle to complete various puzzle tasks that require purely logical reasoning skills rather than professional knowledge, which are easy and even obvious for human~\cite{chollet2019measure, lin2025zebralogic,ma2024kor,wu2025phd}. 
Most existing works targeting puzzles mainly focus on designing benchmarks for evaluation~\cite{borazjanizadeh2024navigating, lin2025zebralogic, wu2025phd}, lacking the training methods and resources for modern LLMs to tackle this challenge.
Existing puzzle datasets often lack diversity and scalability, covering limited puzzle types and offering little control over generation or difficulty~\cite{wu2025phd, lin2025zebralogic, ma2024kor}.
Few works target solving puzzles, yet with either designing a prompting workflow and relying on a code interpreter~\cite{mittal2025fcorebenchlargelanguagemodels}, or training LLMs upon one or a few puzzles~\cite{xie2025logic, tinyzero}, which is difficult to generalize.

Based on the success of the ``LLM{+}RLVR'' paradigm, it has become crucial to obtain a large, diverse, and challenging set of verifiable puzzle prompts for training agents to master logical reasoning.  
To meet this need, we introduce \method—a comprehensive suite that pairs a scalable, controllable puzzle dataset with a training recipe that equips LLMs with strong puzzle-solving abilities.
\method delivers fully synthesizable data across multiple categories, difficulty levels, and scales with automatic evaluation, thus offering effortless integration into RL pipelines.

The whole \method suite comprises of \method-Data, \method-Eval, and \method-Model.
First, \method-Data comprises 36 task types spanning seven broad categories, each probing a distinct facet of logical reasoning.  
Every task is backed by an automated \textit{generator–verifier} pair: the generator can vary both the quantity and the difficulty of the puzzles it produces, while the verifier instantly checks the correctness.
This design yields three key benefits:
1) It can generate unlimited self-verifying puzzle prompts, which plug seamlessly into the RLVR framework and support long chain-of-thought training.  
2) Programmatic difficulty control allows researchers to mix puzzles in desired difficulty ratios and to conduct fine-grained experiments on how curriculum design influences reinforcement learning.  
3) Generators can emit arbitrary sample counts per task, enabling task balancing and cross-task generalization studies.

Based on \method-Data, we present the \method-Eval benchmark, a diverse collection of puzzles that challenges even state-of-the-art LRMs such as o4-mini-high, providing a comprehensive assessment of logical reasoning capabilities.
We introduce the \method-Model recipe, a systematic approach for training high-performance LLMs on puzzle tasks.
This recipe incorporates extensive research on multi-task training strategies, examining various training paradigms and data mixing techniques for controlling data amount and difficulty.
Built upon Qwen2.5-32B~\cite{qwen2.5}, our resulting models surpass current state-of-the-art LRMs on both the \method-Eval benchmark and the challenging ARC-AGI benchmark, while exhibiting remarkable improvements on out-of-domain (OOD) puzzle tasks and maintaining robust generalization in mathematical reasoning.
When scaling the base model from Qwen2.5 (32B dense model) to Seed1.5-Thinking~\cite{seed2025seed} (20B/200B Mixture-of-Expert (MoE) model), we observe a nice generalization from knowledge-orthogonal puzzle data from \method to tasks that require advanced math and STEM domain knowledge and reasoning skills.
Such improvement over a SoTA model seems like a nice ``free lunch'', since the puzzle data are mostly synthetic.

Overall, our key contributions are:
\begin{itemize}
    \item We introduce \method, the first suite for enabling LLMs with advanced and comprehensive logical reasoning abilities for solving puzzles.
    \item The \method suite consists of \method-Data, featuring 36 distinct tasks across 7 categories with controllable difficulty, scalable generation, and automatic verification, which seamlessly fit the RLVR training paradigm.
    \item We establish \method-Eval, a benchmark that rigorously and comprehensively evaluates puzzle reasoning abilities, and propose the \method-Model recipe that trains models with superior performance on in-domain and OOD puzzle reasoning tasks.
\end{itemize}

\section{Related Work}

\paragraph{\textbf{Reinforcement Learning with Verifiable Rewards.}}
Reinforcement Learning (RL) has become a key method for improving models' reasoning capabilities. Unlike Reinforcement Learning with Human Feedback (RLHF), Reinforcement Learning with Verifiable Rewards (RLVR) removes the need for a reward model by directly assigning rewards based on objectively verifiable answers~\cite{guo2025deepseek, seed2025seed, team2025kimi}, which has shown strong performance in mathematics~\cite{yuan2025vapo, yu2025dapo, liu2025understanding}, STEM~\cite{guo2025deepseek, hu2025open}, and coding~\cite{team2025kimi, seed2025seed}. For example, mathematical models are rewarded when their answers match standard solutions~\cite{yu2025dapo, hu2025open}, while coding models receive rewards when their code passes unit tests~\cite{liu2025code}. This method is appealing due to its automation and resistance to reward hacking.
Puzzles are particularly well-suited for RLVR. They can be generated automatically without expert annotations~\cite{albalak2025big, he2025deepmath, xie2025logic}, follow clear rules with algorithmically verifiable solutions~\cite{ma2024kor}, and support precise control over difficulty~\cite{xie2025logic, lin2025zebralogic}. However, most prior RLVR research has focused on other domains, overlooking puzzles' potential for delivering effective reward signals.

\paragraph{\textbf{Puzzle Reasoning of LLMs.}}
Unlike knowledge-intensive tasks, puzzles test a model's reasoning abilities rather than its knowledge or memory capacity~\cite{chollet2019measure, ma2024kor, wu2025phd,chen2025finereason}. Researchers have proposed various benchmarks to assess different types of reasoning, including abstract~\cite{chollet2019measure, wu2025phd}, deductive~\cite{lin2025zebralogic}, and compositional reasoning~\cite{borazjanizadeh2024navigating}. Some benchmarks support scalable generation and difficulty control but lack puzzle diversity~\cite{ wu2025phd, lin2025zebralogic}, while others offer diverse formats without controllable difficulty~\cite{ma2024kor}. A comprehensive benchmark that balances diversity, scalability, and controllability is still missing.
Efforts to improve LLMs' puzzle-solving abilities mainly fall into two categories: tool integration and RLVR. Tool-based methods~\cite{mittal2025fcorebenchlargelanguagemodels} incorporate external resources such as code or symbolic solvers but do not directly enhance the model's internal reasoning. Recent RLVR approaches leverage the scalable generation and verification properties of puzzles~\cite{tinyzero,xie2025logic}, but often focus on a single task type, such as countdown~\cite{tinyzero} or zebra logic~\cite{xie2025logic}, limiting generalization. Our work addresses these gaps by introducing diverse, controllable puzzle datasets and systematically analyzing factors that affect puzzle reasoning performance, including data mixing strategies and multi-task balancing.

\section{\method-Data: The Puzzle Dataset}

In this section, we introduce \method-Data, a comprehensive dataset designed to enhance and evaluate the complex reasoning capabilities of LLMs in RLVR training. 

\subsection{Puzzle Categories}\label{sec:category}
As shown in Figure~\ref{fig:data}, the \method-Data is composed of 36 puzzle tasks of 7 primary categories, including:
\begin{itemize}

    \item \textbf{Crypto Puzzle} is designed to evaluate models' understanding of cryptography and pattern recognition. Tasks such as \texttt{Crypto KPA} and \texttt{Crypto KKA} require models to decode encrypted messages or solve cryptographic challenges, testing their ability to work with hidden or encoded information.
    
    \item \textbf{Arithmetic Puzzle} challenges models to solve problems that require numerical reasoning and basic arithmetic operations. Puzzles like \texttt{Game24} and \texttt{Countdown} test a model's ability to perform arithmetic calculations under constraints.

    \item \textbf{Logic Puzzle} assesses deductive reasoning and the ability to infer conclusions based on premises. Puzzles like \texttt{Knights and Knaves} and \texttt{Zebra Logic} test a model's logical thinking through challenging scenarios that require applying logical rules to solve problems.
    
    \item \textbf{Grid Puzzle} includes tasks that challenge models to solve problems involving structured grids. Puzzles in this category, such as \texttt{Sudoku} and \texttt{Star Battle}, require models to reason about numbers, patterns, and placements in a grid format, testing logical and spatial reasoning abilities.

    \item \textbf{Graph Puzzle} involves tasks where models must reason about nodes, edges, and paths within graph structures. Challenges such as \texttt{Hamiltonian Path} and \texttt{NL Navigation} test a model's ability to understand and traverse graphs, evaluating its capacity for path-finding and network navigation.

    \item \textbf{Search Puzzle} includes tasks that require models to explore a state space efficiently to find a correct solution under specific rules and constraints. Puzzles in this category, such as \texttt{Minesweeper} and \texttt{Tic Tac Toe}, challenge models to simulate or search through potential action sequences, evaluate game or puzzle states, and make optimal decisions. These tasks emphasize planning, local and global search, and reasoning under uncertainty.

    \item \textbf{Sequential Puzzle} focuses on tasks that involve understanding and predicting sequences of steps. Examples include \texttt{Eight Puzzle} and \texttt{Fifteen Puzzle}, which test models' abilities to manipulate objects in a sequence or follow a series of logical steps to reach a solution.
    
\end{itemize}

\subsection{Data Construction}\label{sec:data_construction}
\begin{wraptable}{r}{0.45\linewidth}
    \centering
    \small
    \caption{Task statistics in \method-Data.}
    \begin{tabular}{lc}
    \toprule
        \textbf{Data Type} & \textbf{Task Number} \\ 
        \midrule
        All Puzzles & 36 \\ 
        Auto-Generated Puzzles & 30 \\ 
        \midrule
        Crypto Puzzle & 2 \\ 
        Arithmetic Puzzle & 2 \\ 
        Logic Puzzle & 3 \\ 
        
        Grid Puzzle & 10 \\ 
        
        Graph Puzzle & 4 \\ 
        Search Puzzle & 7 \\ 
        Sequential Puzzle & 8 \\ 
        
        \bottomrule
    \end{tabular}
    
    \label{tab:statistics}
\end{wraptable}
We outline our three-phase data construction pipeline for \method-Data: (1) Tasks Collection and Design; (2) Auto-Generator and Verifier Development; (3) Sliding Difficulty Control. 

\textbf{Phase I: Tasks Collection and Design.}
First, we curate and design 36 logic puzzle tasks that demand complex reasoning capabilities. 
Among these, 30 tasks are scalable with custom generators for creating additional puzzle 
instances, while the remaining 6 tasks draw puzzle instances from existing datasets. 
Most tasks feature multi-step reasoning that integrates various complex reasoning skills. For example, "Twiddle" simultaneously tests spatial manipulation and combinatorial planning abilities.

\textbf{Phase II: Auto-Generator and Verifier Development.} 
Second, we equip 30 tasks in \method with custom puzzle instance auto-generators, which enable automatic data scaling to generate training and evaluation data tailored to o1-like complex reasoning research.
In addition, all 36 puzzle tasks have corresponding auto-verifiers that undergo manual validation. 
They evaluate the correctness of models' outputs or provide outcome rewards and penalty scores for complete reasoning chains.

\textbf{Phase III: Sliding Difficulty Control.}
For each puzzle task, we identify key variables that control difficulty, such as grid size and blank cell count in \texttt{Binario}. These variables serve as parameters in our auto-generator to create puzzle instances across difficulty levels.
We evaluate model performance on these puzzles using the \textit{pass@k}$=\mathbb{E}_{\text{Problems}} \left[ 1 - \frac{\binom{n - c}{k}}{\binom{n}{k}} \right]$ metric ($n=200; k=1, 10, 100$) following~\cite{chen2021evaluating}. We establish three difficulty levels (Easy, Medium, Hard) for each task by analyzing performance trends across different parameter settings. 

\subsection{Task Statistics}
Table~\ref{tab:comparison} presents a comparison between \method and existing puzzle resources. \method is the only dataset encompassing multiple task categories, offers scalability, provides automatic verification, and is publicly available. Additionally, it uniquely employs the RLVR approach to enhance models' puzzle reasoning capabilities fundamentally.
Table~\ref{tab:statistics} presents the distribution of tasks across the \method dataset.

\begin{table*}[t]
    \centering
    \small
    \caption{Comparison of different puzzle reasoning benchmarks.}
    \begin{tabular}{lcccccc}
    \toprule
        \textbf{Resources} & \textbf{Categories} & \textbf{Tasks} & \textbf{Scalable} & \textbf{Auto-Verifier} & \textbf{Trainable} & \textbf{Solution} \\ 
        \midrule
        KOR-Bench~\cite{ma2024kor} & 5 & 125 & \ding{55}  & \ding{51} & \ding{55} & N/A\\ 
        NPR~\cite{wu2025phd} & 1 & 1 & \ding{55} & \ding{55} & \ding{55} & N/A\\ 
        ZebraLogic~\cite{lin2025zebralogic} & 1 & 1 & \ding{51} & \ding{51} & \ding{51} & N/A\\ 
        SearchBench~\cite{borazjanizadeh2024navigating} & 1 & 11 & \ding{51} & \ding{51} & \ding{55} & N/A\\ 
        FCoReBench~\cite{mittal2025fcorebenchlargelanguagemodels} & 5 & 40 & \ding{51} & \ding{51} & \ding{55} & Prompt-based\\ 
        Logic-RL~\cite{xie2025logic} & 1 & 1 & \ding{51} & \ding{51} & \ding{51} & RLVR\\
        \method & 7 & 36 & \ding{51} & \ding{51} & \ding{51} & RLVR\\ 
        \bottomrule
    \end{tabular}
    \label{tab:comparison}
\end{table*}

\subsection{\method-Eval}\label{sec:eval_data}
We develop \method-Eval by systematically sampling from our broader dataset. For each task, we aimed to extract 50 instances per difficulty level (Easy, Medium, Hard). However, due to inherent constraints in some tasks, we collected 4,758 puzzle instances rather than the theoretical maximum of 5,400. This discrepancy arises because some tasks generate fewer than 50 instances per difficulty level, while others rely on manually collected and annotated data rather than auto-generation. Importantly, we ensured no data leakage between training and evaluation sets by implementing strict separation protocols during the sampling process.

\section{\method-Model: The Training Recipe}

Developing advanced logical reasoning in language models requires a carefully structured training approach that develops diverse reasoning skills while avoiding overfitting specific problem types.
Our training methodology follows a two-stage process designed to build reasoning abilities systematically: 
(1) rejection fine-tuning to establish foundational reasoning patterns and
(2) multi-task RL to develop general reasoning skills that transfer across diverse problem domains.

\subsection{Rejection Fine-tuning}
Directly applying reinforcement learning to a base model often results in training instability and may not unlock the model's full performance potential~\cite{guo2025deepseek}.
To address this, we begin with rejection fine-tuning (RFT), i.e., leveraging high-quality solutions during supervised fine-tuning (SFT) to establish solid foundational reasoning patterns.
In our training data, we strategically combine math problems with puzzles, as mathematics elicits diverse reasoning patterns and contributes to the model generalization~\cite{shen2025exploring}.

For puzzles, we uniformly sample tasks and difficulty levels from the \method dataset to ensure a comprehensive coverage and balanced distribution of reasoning patterns.
We also include the training data of the ARC-AGI puzzle~\cite{chollet2019measure, arcagi2} in the RFT data, since it is too difficult to learn without RFT as cold-start.
For each puzzle instance, we utilize DeepSeek-R1~\cite{guo2025deepseek} to generate 8 candidate solutions, from which we select the correct solution for RFT.
The mathematical component consists of carefully curated examples from a high-quality R1-distilled mathematical dataset~\cite{wen2025light}. 
Throughout RFT, we maintain a balanced 1:1 ratio between puzzles and mathematical problems to ensure comprehensive reasoning development across domains.
Detailed implementation and dataset specifications are provided in Appendices~\ref{appendix:datadetails} and~\ref{appendix:imdetails}.

\subsection{RL with Verifiable Puzzles}\label{sec:RLVR}
We use VC-PPO~\cite{yuan2025s}, a PPO variant~\cite{schulman2017proximal}, to train our models.  
Each of the 36 tasks in \method has an automated verifier $v_i$ that instantly scores a response as correct or incorrect.  
For 30 tasks we also have a generator $g_i$ that can create examples at any difficulty; the other 6 tasks draw from fixed pools $F_i$.
For each task $i$ and difficulty level $d\in\mathcal D_i$, we choose how many examples $N_{i,d}$ to use. Then:
\[
s_i = 
\begin{cases}
  \displaystyle\sum_{d\in\mathcal D_i} g_i\bigl(N_{i,d},d\bigr), 
  & i\le 30,\\[1ex]
  \displaystyle\sum_{d\in\mathcal D_i}\mathrm{Sample}\bigl(F_i^d,\min(N_{i,d},|F_i^d|)\bigr),
  & i>30.
\end{cases}
\]
The full training set is $S=\bigcup_i s_i$, with total size $|S|=\sum_i|s_i|$.  By changing the $N_{i,d}$, we can easily adjust:
1) How many examples come from each task,
2) The mix of easy vs.\ hard items,
3) Overall dataset size.
During training, each generated example is fed to its verifier $v_i$, which returns a reward that VC-PPO uses to update the policy. This loop provides a fully automatic RL pipeline for puzzle reasoning.

\subsection{Multi-task Training}
\label{sec:multi-task}
Developing general logical reasoning is hard because different puzzles require different thinking skills. 
To build strong, transferable problem-solving abilities, we explore two multi-task training methods: \textbf{Mix-training RL} and \textbf{Multi-stage RL}, since single-task training often leads to narrow expertise and poor transfer to new puzzles~\cite{xie2025logic}.

A rich diversity of tasks can significantly enhance generalization and actively prevent overspecialization~\cite{guo2025deepseek}. 
Therefore, we employ \textbf{Mix-training RL} to integrate multiple puzzle types simultaneously during the training process.
Our methodology involves a meticulously constructed dataset that integrates three critical components: 
a) The training split of \method, featuring balanced task and difficulty distributions; 
b) The public training set of ARC-AGI 1 and 2, which improve the generalization of existing reasoning abilities to unseen tasks; and 
c) AIME mathematical problems (1983-2023), which are difficult enough to elicit diverse reasoning patterns and enhance generalization.
A strategic 1:1 puzzle-to-mathematics ratio is maintained throughout this training process to foster the development of complementary reasoning systems within the model.

Mix-training RL offers broad exposure to different puzzle types, but the varied reasoning skills required can cause conflicts between tasks.
To address this, we adopt Multi-stage RL, a curriculum-based approach that builds core skills before introducing new challenges.
For difficult tasks like ARC-AGI, we use a two-phase strategy:
1) train intensively on ARC-AGI 1, 2, and AIME until the model generalizes well and performance stabilizes;
2) gradually introduce \method-Data while retaining earlier data to avoid forgetting.
This step-by-step method helps the model learn complex reasoning more effectively and maintain strong performance on earlier tasks.
More implementation details are provided in Appendix~\ref{appendix:datadetails}.

\section{Experiments}

\label{sec:experiment}
\subsection{Experiment Setup}
We adopt several challenging reasoning benchmarks for evaluation: \method-Eval, and abstract reasoning challenges ARC-AGI 1~\cite{chollet2019measure} and ARC-AGI 2~\cite{arcagi2} known for their extreme difficulty for LLMs.
We also include the knowledge-orthogonal reasoning benchmark KOR-Bench~\cite{ma2024kor}, which contains puzzles from five categories.
To examine generalization capabilities, we include the advanced mathematics benchmark AIME 2024.
We evaluate each AIME problem 32 times and others 4 times, reporting the average performance for reliability.
Baselines are described in Appendix~\ref{appendix:baselines}.

We train our models from Qwen2.5-32B-Instruct~\cite{qwen2.5}, a solid starting point for training strong reasoning models~\cite{qwq32b,guo2025deepseek}.
After acquiring the RFT model (Qwen2.5-32B-RFT), we leverage the Mix-Training approach with 370 training steps to get the RL model (Qwen2.5-32B-\method), showing superior overall performance in our experiments.
Each PPO step performs multiple gradient updates by iterating over 4 mini-batches derived from the training batch.
Details are described in Appendix~\ref{appendix:imdetails}.

\subsection{Results} 
\begin{table*}[t]
    \caption{Performance of reasoning models, generic models, and our trained LLMs on reasoning benchmarks.}
    \centering
\small
\begin{tabular}{lccccc}
\toprule
\multirow{3}{*}{\textbf{Model}} &
  \multicolumn{4}{c}{\textbf{Puzzle}} &
  \textbf{Math} \\
\cmidrule(lr){2-5} \cmidrule(lr){6-6}

 & 
  \multicolumn{3}{c}{\textbf{In-Domain}} &
  \multicolumn{1}{c}{\textbf{Out-of-Domain}} &
  \multirow{2}{*}{\textbf{AIME 24}} 
  \\
  \cmidrule(lr){2-4} \cmidrule(lr){5-5} 
 &
  \textbf{ARC-AGI 1} &
  \textbf{ARC-AGI 2} &
  \textbf{\method-Eval} &
  \textbf{KORBench} &
   \\
\midrule
o4-mini-high               & 54.7 & 2.6 & 65.1 & 72.7 & 93.4 \\
o3-mini-high               & 25.8 & 0.4 & 59.9 & 69.6 & 87.3 \\
o1                  & 29.0 & 0.4 & 54.9 & 69.9 & 74.3 \\
DeepSeek-R1                & 17.8 & 0.2 & 49.2 & 71.7 & 79.8 \\
Gemini-2.5-Pro             & 22.7 & 1.4 & 50.6 & 68.2 & 90.3 \\

Claude-3.7-Sonnet-Thinking & 37.6 & 1.4 & 53.2 & 67.8 & 60.3 \\
DS-R1-Distilled-Qwen-32B   & 7.9  & 0.0 & 31.1 & 63.5 & 72.0 \\
QwQ-32B & 7.0 &	0.0  & 43.8  &  61.8  & 69.2 \\

\cmidrule(lr){1-6}

Grok-2-1212                & 7.5  & 0.0 & 13.6 & 54.9 & 16.7 \\
GPT-4o-1120                & 7.3  & 0.0 & 14.2 & 57.9 & 10.0 \\
Claude-3.7-Sonnet          & 12.9 & 0.0 & 22.7 & 56.9 & 23.0 \\
\cmidrule(lr){1-6}

Qwen2.5-32B-Instruct          & 6.0 & 0.0 & 12.6 & 54.7 & 16.6 \\
Qwen2.5-32B-RFT & 8.4&0.0 & 46.6& 61.0& 62.0\\
Qwen2.5-32B-\method          & 32.8 & 0.6 & 62.6 &  65.0 & 60.6 \\

\bottomrule
\end{tabular}
    \label{tab:040overall}
\end{table*}

According to Table~\ref{tab:040overall}, our model outperforms most of the public models on \method-Eval with 32B parameters, demonstrating the effectiveness of our dataset and training recipe. 
Besides, our model stands out on the challenging ARC-AGI benchmark, surpassing strong reasoning models such as Gemini 2.5 Pro, o3-mini, and o1. 
Additionally, RFT and multi-task RL training strategies yield significant performance gains on OOD benchmarks.
This indicates that our training recipe effectively enhances the model's general logic reasoning abilities and can generalize to unseen tasks.
Moreover, after RL training, our models maintain comparable math reasoning abilities gained from rejection fine-tuning, showing that our training strategy preserves general reasoning capabilities while enhancing logic-specific skills.

\begin{table*}[t]
    \caption{Performance of reasoning LLMs, generic LLMs, and our trained LLMs on \method-Eval.}
    \centering
\small
    \begin{tabular}{l c ccccccc}
    \toprule
    \textbf{Model} & 
    \textbf{Crypto} &
    \textbf{Arithmetic} &
    \textbf{Logic}  &
    \textbf{Grid}  &
    \textbf{Graph}  &
    \textbf{Search} &
    \textbf{Sequential}  &
    \textbf{Overall}\\
    \midrule

    o4-mini-high & 97.0 & 93.3 & 82.2 & 71.0 & 62.4 & 64.3 & 34.0 &  \textbf{65.1} \\
    
    o3-mini-high & 88.1 & 76.9 & 74.3 & 65.5 & 63.1 & 61.5 & 29.6  &  \textbf{59.9} \\
    
    o1 & 97.8 & 80.0 & 60.1 & 63.3 & 56.2 & 50.9 & 23.6 &  \textbf{54.9} \\
    
    DeepSeek-R1 & 82.7 & 77.1 & 71.4 & 51.1 & 62.6 & 38.4 & 19.5 &  \textbf{49.2} \\

    Gemini-2.5-Pro & 75.2 & 95.4 & 71.5 & 58.9 & 37.3 & 49.4 & 17.5 &  \textbf{50.6} \\

    Claude-3.7-Sonnet-Thinking & 81.5 & 75.4 & 76.6 & 57.7 & 50.5 & 49.4 & 26.4 &  \textbf{53.2} \\
    
    DS-R1-Distilled-Qwen-32B & 16.7 & 47.3 & 62.9  & 38.2 & 36.4 & 12.4 & 18.8 &  \textbf{31.1}  \\

    QwQ-32B &	59.0 	&65.7 	 &74.4 	&47.5 	&47.3 	&28.2 	&24.4 & \textbf{43.8} \\
        
    \cmidrule(lr){1-9}
    Grok-2-1212 & 10.1 & 9.4 & 50.0 & 12.8 & 17.6 & 3.9 & 6.4 &  \textbf{13.6} \\
    
    GPT-4o-1120 & 26.2 & 1.9 & 34.5 & 17.8 & 19.3 & 6.0 & 3.9 &  \textbf{14.2} \\

    Claude-3.7-Sonnet & 38.1 & 16.7 & 60.0 & 22.9 & 22.4 & 7.8 & 15.0 &  \textbf{22.7 }\\

    \cmidrule(lr){1-9}

    Qwen2.5-32B-Instruct &	4.0 &	10.3 	&46.4 	&15.3 &	7.3 &	2.5 	&8.2 &	 \textbf{12.6} \\

    Qwen2.5-32B-RFT & 62.0&71.7 & 71.6& 51.3& 55.1& 39.3& 18.4& \textbf{46.6}\\
    Qwen2.5-32B-\method & 96.0 & 93.7 & 90.2 & 62.6 & 54.0 & 70.4 & 29.7 &  \textbf{62.6} \\
    
    \bottomrule
    \end{tabular}

    \label{tab:040apeval}
\end{table*}

Next, let's dive into detailed analysis across reasoning categories in \method-Eval.
In Table~\ref{tab:040apeval}, Qwen2.5-32B-\method demonstrates exceptional performance in structured reasoning categories, particularly excelling in Crypto, Arithmetic, and Logic tasks. 
This suggests that our training approach effectively develops capabilities in rule-based reasoning with explicit constraints and patterns.
Besides, our model shows competitive performance in search tasks, outperforming most baseline models. Search problems require strategic exploration of solution spaces and planning capabilities. The strong performance suggests that our approach effectively develops these higher-order reasoning skills.
Notably, we observe a consistent performance hierarchy across categories for most models. 
Crypto and Arithmetic tasks tend to yield the highest accuracy, while spatial and sequential tasks remain more difficult. 
These challenges point to promising directions for future work.

\subsection{Generalization with Scaling: Free Lunch from \method}

\begin{table}[t]
    \centering
    \small
    \caption{Results on benchmarks for general reasoning capabilities. This demonstrates that additional RL training on \method-Data generalizes well on larger models, showing the benefits of puzzle data for math and STEM reasoning.}
    \begin{tabular}{lcccc}
    \toprule
        \textbf{Model} & \textbf{AIME 2024} & \textbf{AIME 2025} & \textbf{BeyondAIME} & \textbf{GPQA Diamond} \\
        \midrule
        o4-mini-high & 93.4 & 92.7 & 55.7 & 81.4 \\
        o3-mini-high & 87.3& 86.5& 63.6& 79.7\\
        o1 & 74.3 & 79.2 & 50.0 & 78.0\\
        DeepSeek-R1 & 79.8 & 65.0 & 42.4 & 71.5  \\
        \midrule
        Seed1.5-Thinking & 86.7 & 74.0 & 48.0 & 77.3  \\
        Seed1.5-Thinking-\method & $87.5_{(+0.8)}$ & $75.9_{(+1.9)}$ & $48.4_{(+0.4)}$ & $78.1_{(+0.8)}$ \\
    \bottomrule
    \end{tabular}
    \label{tab:seed1.5t}
\end{table}

Can logical reasoning data, such as puzzles that do not require domain knowledge, benefit general reasoning capabilities?
We do not observe such a generalization for Qwen2.5-32B. 
Therefore, we scale the experiments to a much larger model.
We follow Seed1.5-Thinking~\cite{seed2025seed} and train from the same Mixture-of-Experts (MoE) model in the RL stage, which features 20B activated and 200B total parameters.
To make a fair comparison between Seed1.5-Thinking, we adopt the same base model (20B/200B) and the same RL training data except for 20K \method-Data, and train the models for comparable PPO steps.

Surprisingly, the results in Table~\ref{tab:seed1.5t} show that our dataset enhances general capabilities like math and STEM problem-solving.  
When compared to Seed1.5-Thinking, a leading reasoning model, additional training with \method, i.e., Seed1.5-Thinking-\method, generally improves performance on AIME 2024 and 2025, BeyondAIME~\cite{seed2025seed} (an expert-curated, more challenging evaluation dataset), and GPQA Diamond~\cite{rein2024gpqa}.
Given the difficulty of further improving SoTA models like Seed1.5-Thinking, simply incorporating \method’s synthetic puzzle data during the RL training stage appears almost like a ``free lunch'' for expanding the capability spectrum of reasoning models, even leading to generalization improvements in general advanced reasoning.\footnote{Note that, we primarily experiment upon Qwen2.5-32B for further analysis due to the prohibitive resources required to train a 20B/200B model.}

\subsection{Ablation Studies} 
\label{sec:ablation}
\paragraph{\textbf{Training Data Size.}}
\begin{figure}[t]
    \centering
    \includegraphics[width=\textwidth]{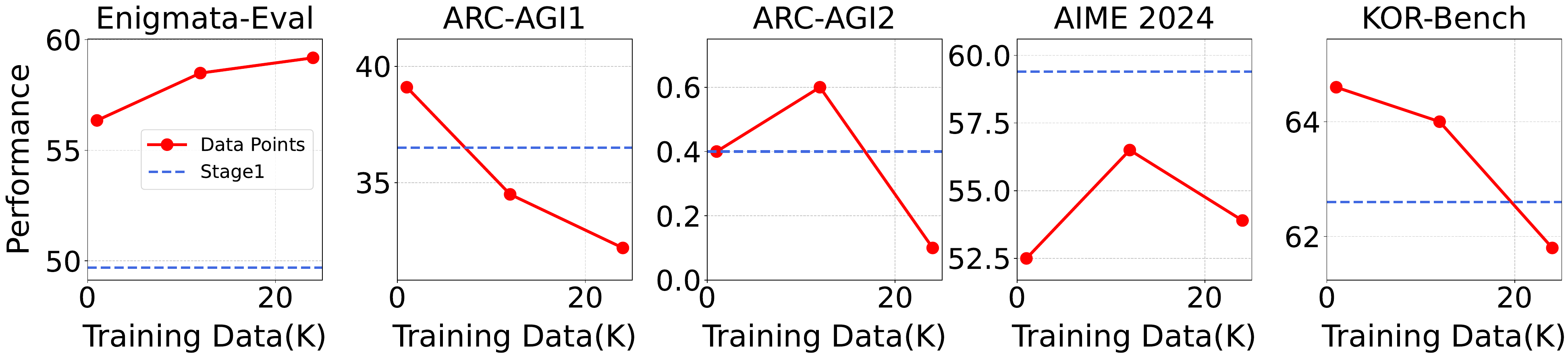}
    \caption{Impact of training data size in the second stage of Multi-stage Training on model performance across different benchmarks. The blue dashed line represents model performance after the first training stage, while the red solid line shows performance after the second stage.}
    \label{fig:data_amount}
\end{figure}

We study the impact of varying sizes of \method-Train data on model performance during the second stage of Multi-stage Training.
To ensure a fair comparison, all checkpoints were evaluated using models obtained at step 150 from the sample stage-1 checkpoint, with equal sampling across all difficulty levels.
As shown in Figure~\ref{fig:data_amount}, first, a small amount of \method-Train data in the second stage significantly improves \method-Eval performance while better preserving first-stage knowledge and OOD performance.
Second, increasing \method-Train data progressively enhances in-domain \method-Eval performance.
Third, excessive \method-Train data leads to catastrophic forgetting and slightly degraded OOD performance.

\begin{table*}[t]
    \caption{Comparison between different data mixing strategies in the stage 2 of Multi-stage RL.}
    \centering
\small
\begin{tabular}{lccccc}
\toprule
\multirow{3}{*}{\textbf{Mixing Method}} &
  \multicolumn{4}{c}{\textbf{Puzzle}} &
  \textbf{Math}  \\
  \cmidrule(lr){2-5} \cmidrule(lr){6-6}
 &
  \multicolumn{3}{c}{\textbf{In-Domain}} &
  \multicolumn{1}{c}{\textbf{Out-of-Domain}} &
  \multirow{2}{*}{\textbf{AIME 24}} \\
  \cmidrule(lr){2-4} \cmidrule(lr){5-5} 
  &
  \textbf{ARC-AGI 1} &
  \textbf{ARC-AGI 2} &
  \textbf{\method-Eval} & 
  \multicolumn{1}{c}{\textbf{KORBench}} &
\\
\midrule
HRV 
   & 34.7
   & 0.1
   & 57.7
   & 60.1
   & 50.1 	
    
   \\
Easy:Medium:Hard = 1:1:1
   & 34.5 	 
   & \textbf{0.6}
   & 58.5 	
   & \textbf{64.0} 
   & \textbf{56.6} 	
    
   \\
Easy:Medium:Hard = 2:6:2
   & \textbf{35.1}
   & 0.3	
   & \textbf{58.9} 	
   & 63.8  
   & 48.0 	 
   	
   \\
  \bottomrule
\end{tabular}

    \label{tab:040mixing}
\end{table*}

\paragraph{\textbf{Data Difficulty Control.}}
We study how the distribution of data difficulty affects performance. 
As described in $\mathsection$~\ref{sec:RLVR}, we set the data size at different difficulty levels as $N_{i}^d$. 
With $N_i = 400$ per task, we compare two ratios in Multi-stage Training's second stage: balanced ($N_{i}^{\text{easy}} : N_{i}^{\text{med}} : N_{i}^{\text{hard}}$=1:1:1) versus medium-focused ($N_{i}^{\text{easy}} : N_{i}^{\text{med}} : N_{i}^{\text{hard}}$ = 2:6:2). 
The latter setting quantifies how extreme samples can undermine RL training~\cite{yu2025dapo}. 
We also compare with historical reward variation (HRV)~\cite{wang2025reinforcement} as baseline data selection strategy, using the same stage-1 checkpoint and 150 second-stage steps.
As shown in Table~\ref{tab:040mixing}, the balanced difficulty ratio (1:1:1) enables the model to demonstrate more robust complex reasoning performance. 
Also, our effortless difficulty control method based on difficulty tags in \method data performs comparably to HRV on \method-Eval while delivering superior results on OOD benchmarks.

\paragraph{\textbf{Multi-task Training.}}
\begin{table*}[t]
    \caption{Comparison between different training strategies.}
    \centering
\small
\begin{tabular}{lccccc}
\toprule
\multirow{3}{*}{\textbf{Model}} &
  \multicolumn{4}{c}{\textbf{Puzzle}} &
  \textbf{Math}  \\
  \cmidrule(lr){2-5} \cmidrule(lr){6-6}
 &
  \multicolumn{3}{c}{\textbf{In-Domain}} &
  \multicolumn{1}{c}{\textbf{Out-of-Domain}} &
  \multirow{2}{*}{\textbf{AIME 24}} \\
  \cmidrule(lr){2-4} \cmidrule(lr){5-5} 
  &
  \textbf{ARC-AGI 1} &
  \textbf{ARC-AGI 2} &
  \textbf{\method-Eval} & 
  \multicolumn{1}{c}{\textbf{KORBench}} &
\\
\midrule
Mix Training + SFT-Part 
   & 22.7 	
   & 0.0 
   & 46.7
   & 56.7
   & 51.8 	
    
   \\
Multi-Stage + SFT-Part	
   & 27.5 	 
   & 0.0 
   & 56.1 	
   & 51.6 
   & 45.9 	
    
   \\
Mix Training + SFT-All
   & \textbf{34.5 }	
   & 0.1 	
   & \textbf{62.6 }	
   & \textbf{60.4 } 
   & \textbf{58.8 	} 
   	
   \\
Multi-Stage + SFT-All
   & 33.4 	
   & \textbf{0.4 }	
   & 61.1
   & 60.2 
   & 55.6 	
   
   \\
  \bottomrule
\end{tabular}

    \label{tab:040MS}
\end{table*}

For the two training paradigms (Multi-stage RL and Mix-Training RL) in $\mathsection$~\ref{sec:multi-task}, we evaluate the impact of SFT/RFT using two variants: 1) \textbf{SFT$_{\texttt{Part}}$}, which excludes three tasks (Countdown, Minesweeper, Light Up) to test transfer ability; and 2) \textbf{SFT$_{\texttt{All}}$}, which includes all tasks as a complete baseline.
To ensure fair comparison, all checkpoints were trained for identical total steps: (1) Multi-Stage training consisting of 200 steps in stage 1 and 225 steps in stage 2; and (2) Mix-Training using the combined dataset across all 425 steps.
As shown in Table~\ref{tab:040MS}, first, Multi-Stage and Mix-Training approaches show complementary strengths. 
Multi-Stage builds deeper task-specific reasoning, while Mix-Training improves generalization.
Second, With limited pre-training data, Multi-Stage transfers better to unseen tasks, especially complex ones like \method and ARC-AGI, reflecting curriculum learning benefits.
Third, Mix-Training generalizes better to OOD tasks, suggesting that diverse training helps models learn broader reasoning strategies beyond specific tasks.

Moreover, Figure~\ref{fig:len_reward} shows Mix-Training RL's and Multi-Stage RL's training dynamics. 
We observe a positive correlation between rewards and response length. Both approaches achieve similar final rewards, showing the signs of test-time scaling effect~\cite{snell2024scaling,zhang2025and} where models learn to explore more tokens to find better solutions. 
However, the more volatile response length under Multi-Stage RL suggests instability in generation. In contrast, Mix-Training RL yields more consistent outputs with smoother length curves, indicating greater training stability.

\subsection{Analysis}

\begin{figure}[t]
    \centering
    \includegraphics[width=\textwidth]{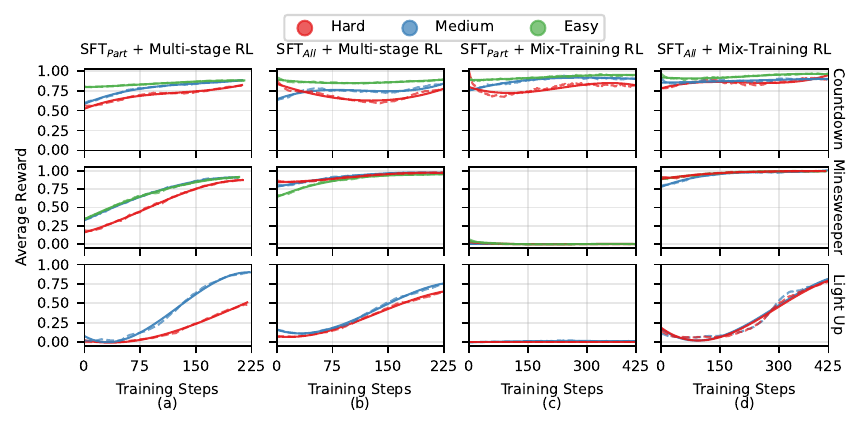}
    \caption{Learning curves across training approaches for representative puzzle tasks. Each row represents a different task, and each column represents a different training approach. The curves show how the average reward changes with training steps for different difficulty levels.}
    \label{fig:multi_task_comparison}
\end{figure}

\paragraph{\textbf{How does SFT affect RL training?}}

We further analyze the influence of SFT on RL training to explore the reason behind the performance gap in Table~\ref{sec:multi-task}. 
We represent the reward curves across different training steps for different training approaches in Figure~\ref{fig:multi_task_comparison}.
For simple tasks like Countdown, all methods achieve similar improvements, suggesting that SFT is not essential. 
For medium-complexity tasks like Minesweeper, Mix-Training struggles without SFT, while Multi-stage RL still learns effectively.
\begin{wraptable}{r}{0.45\linewidth}
    \caption{Impact of code utilization on Qwen2.5-32B-\method accuracy}
    \centering
    \small
    \begin{tabular}{lcc}
        \toprule
        \textbf{Condition} & \textbf{ARC-AGI 1} & \textbf{\method} \\
        \midrule
        Overall & 34.5 & 62.6\\
        With code & 12.5 & 41.8\\
        Without code & \textbf{35.7} & \textbf{63.6}\\
        \bottomrule
    \end{tabular}
    \label{tab:code_utilization}
\end{wraptable}
When the task is included in SFT$_{\text{All}}$, both approaches start from a higher baseline and quickly optimize to near-perfect accuracy. 
This pattern is typical for tasks with fixed solution patterns or shortcuts—once the model grasps the solving approach, performance improves dramatically.
For high-complexity tasks like Light Up, the combination of comprehensive SFT and Multi-stage RL dramatically outperforms all other approaches, particularly for difficult variants of the task. 
Interestingly, Mix-Training RL without relevant SFT (c) fails, highlighting that complex reasoning tasks require both strong foundational knowledge (from relevant SFT) and structured learning approaches (from Multi-stage RL) to achieve optimal results.

\begin{figure}[t]
    \centering
    \begin{minipage}[b]{0.33\linewidth}
        \includegraphics[width=\linewidth]{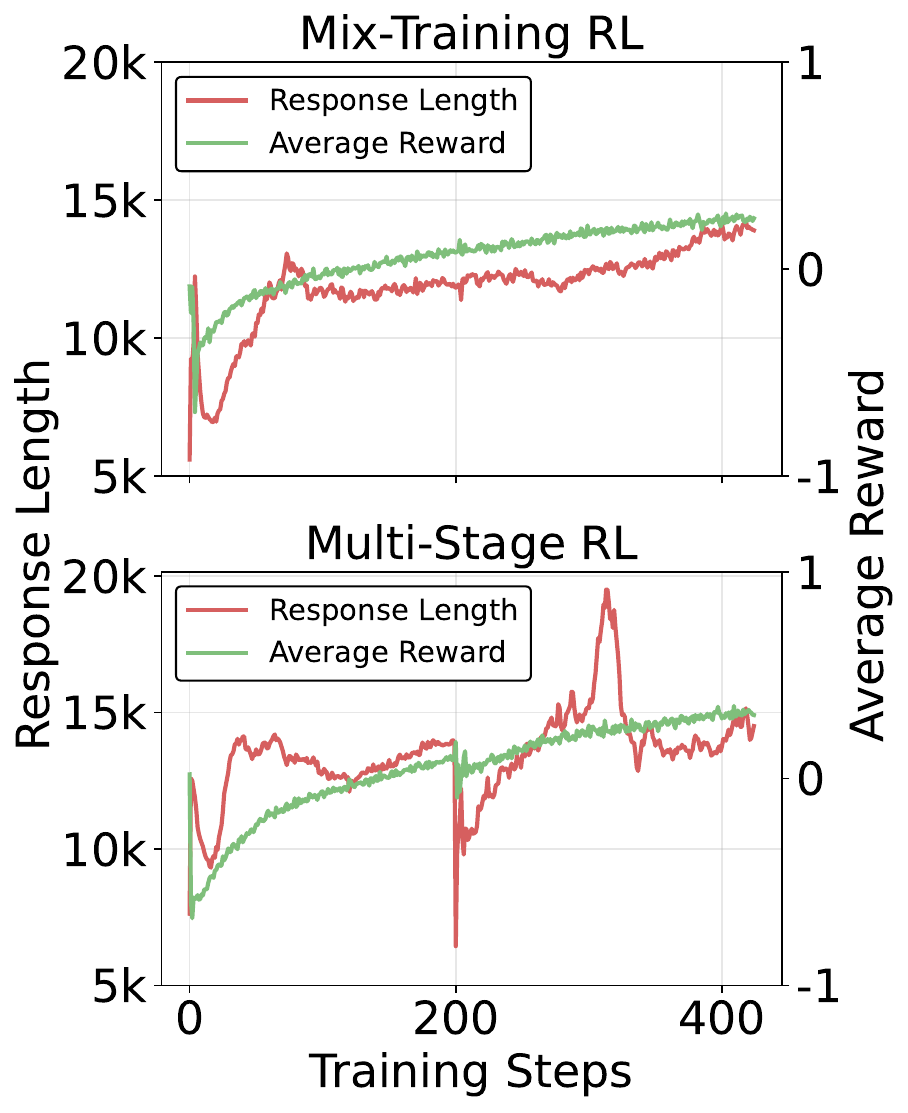}
    \caption{The response length and reward curves during Mix-Training RL and Multi-Stage RL training.}
    \label{fig:len_reward}
    \end{minipage}
    \hfill 
    \begin{minipage}[b]{0.65\linewidth}
        \includegraphics[width=\linewidth]{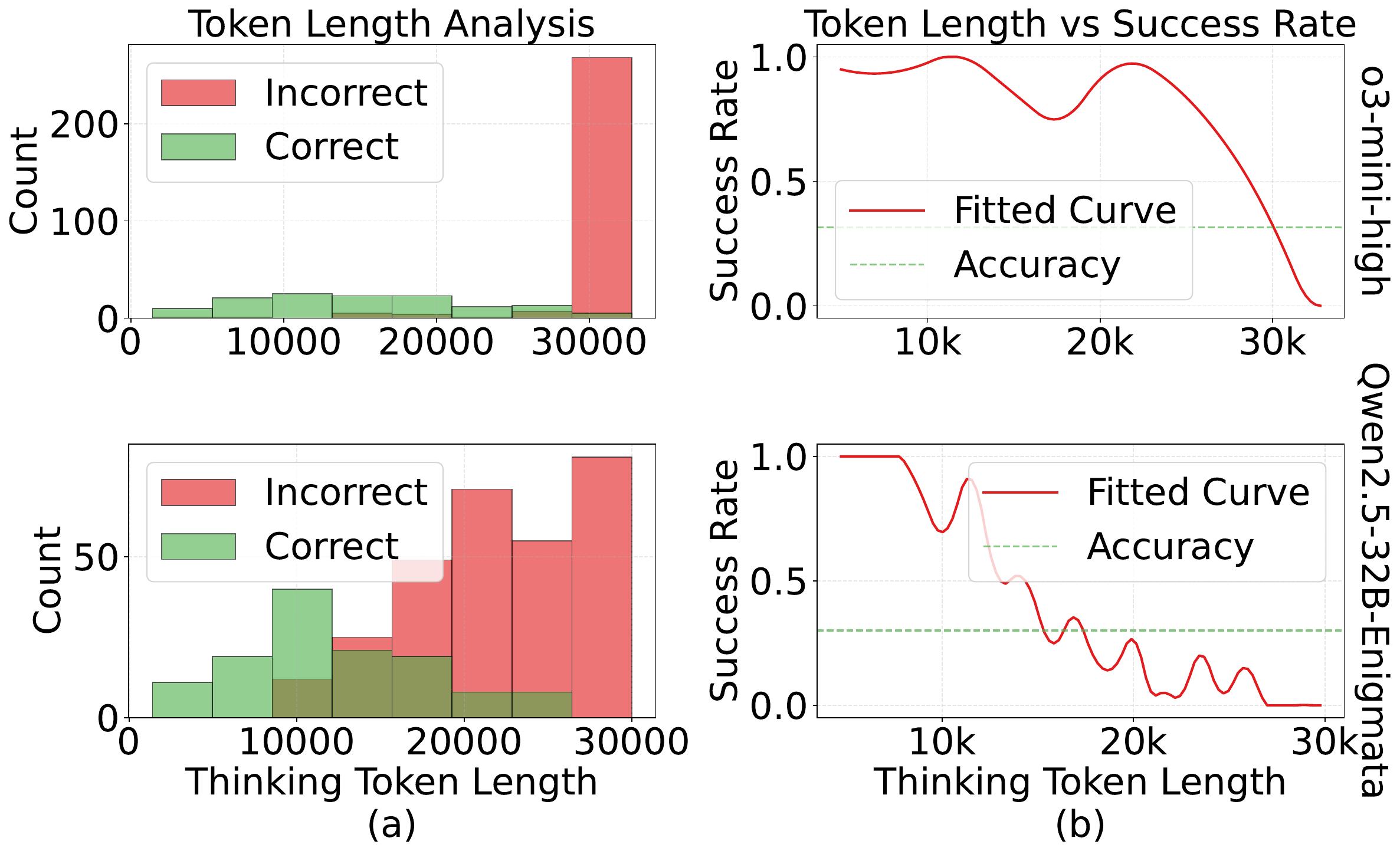}
    \caption{Performance analysis of Qwen2.5-32B-\method on ARC-AGI 1: 
    (a) distribution of reasoning token lengths for correct and incorrect responses, 
    (b) and success rate by reasoning token length.}
    \label{fig:output_analysis}
    \end{minipage}
\end{figure}

\paragraph{\textbf{Response Length Analysis.}}

Figure~\ref{fig:output_analysis} presents the distribution of reasoning token lengths for both correct and incorrect responses in ARC-AGI 1. 
Tasks, where the model produces more concise reasoning, showed significantly higher accuracy than those requiring longer reasoning chains. 
This analysis identifies potential inefficiencies in the model's reasoning process and highlights opportunities for reasoning optimization.

\paragraph{\textbf{Code Utilization in Puzzle Reasoning.}}

We investigated whether code utilization enhances model performance on reasoning tasks. We identified code elements in model outputs using keyword detection and pattern matching and classified responses accordingly. 
According to Table~\ref{tab:code_utilization}, code utilization hindered puzzle task performance. 
This suggests that current models fail to effectively use code \textit{without executing} it for complex reasoning tasks.

\section{Conclusion}

In this paper, we present \method, a suite for equipping LLMs with advanced puzzle reasoning, which integrates seamlessly with reinforcement learning using verifiable rule-based rewards. 
\method-Data features 36 tasks across seven reasoning categories, with its scalable generation, automated verification, and adjustable difficulty.
We also introduce the \method-Eval benchmark for assessing puzzle reasoning abilities and guiding research on generalizable reasoning models.
\method-Model, trained with RLVR, demonstrates its superior performance and robust generalization and reasoning skills.
Experiments also demonstrate that when trained on larger models such as Seed1.5-Thinking (20B/200B), synthetic puzzle data brings extra benefits in other domains, such as math and STEM reasoning, over a SoTA model.
We hope \method serves as a solid foundation for the community to push forth the research on reasoning models.

\newpage
\section{Contributions}
\label{sec:contributions}

\subsection*{Project Lead}

Jiangjie Chen$^{1,6}$

\subsection*{Core Contributors\footnote{Equal Contribution. Order decided by institutions.}}
Jiangjie Chen$^{1,6}$, Qianyu He$^{1,2}$\footnote{Work done during internship at ByteDance Seed.},
Siyu Yuan$^{1,2}$\footnotemark[3],
Aili Chen$^{2}$

\subsection*{Contributors}
Zhicheng Cai$^{4,6}$, Weinan Dai$^{1,3,6}$\footnotemark[3], Hongli Yu$^{1,3,6}$\footnotemark[3], Qiying Yu$^{1,3,6}$\footnotemark[3], Xuefeng Li$^{1,5}$\footnotemark[3], Jiaze Chen$^{1}$

\subsection*{Supervisors}
Hao Zhou$^{3,6}$, Mingxuan Wang$^{1,6}$

\subsection*{Affiliation}

$^1$ByteDance Seed

$^2$Fudan University

$^3$Institute for AI Industry Research (AIR), Tsinghua University

$^4$Nanjing University

$^5$Shanghai Jiao Tong University

$^6$SIA-Lab of Tsinghua AIR and ByteDance Seed

\clearpage

\bibliographystyle{unsrt}
\bibliography{main}

\clearpage

\beginappendix

\section{Reward Curves Across Individual Tasks in \method}
\label{appendix:reward_curves}

\begin{figure}[t]
    \centering
    \includegraphics[width=0.8\textwidth]{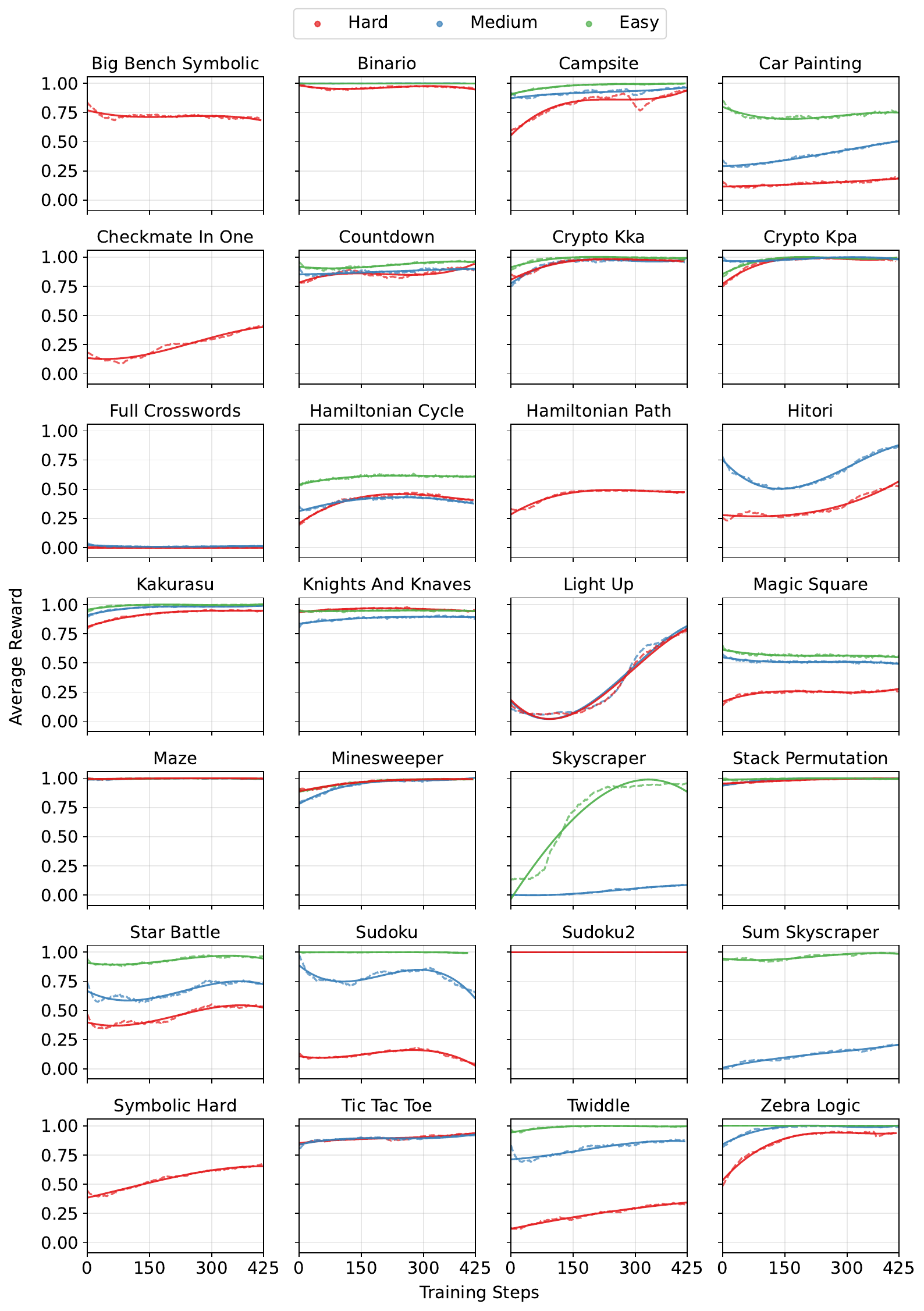}
    \caption{Reward curves for Qwen2.5-32B-\method across all individual tasks during training. Each subplot represents a different puzzle task, with the x-axis showing training steps and the y-axis showing average reward. Colors indicate different difficulty levels: Easy (green), Medium (blue), and Hard (red).}
    \label{fig:individual_task_rewards}
\end{figure}

We conducted a detailed analysis of our model's learning dynamics by examining reward curves across all tasks in the \method dataset. Figure~\ref{fig:individual_task_rewards} presents these learning trajectories, revealing several important patterns in how models acquire puzzle-solving capabilities.

The reward curves across tasks reveal three distinct learning patterns:

\textbf{1. Gradual Mastery Tasks:}
Several tasks, such as \texttt{Light Up} and \texttt{Zebra Logic} show smooth and consistent reward gains throughout training. These tasks typically involve complex reasoning chains or require integrating multiple constraints, making them suitable for long-horizon learning. The steady upward trend suggests that the agent is progressively refining its decision-making strategy through extended exploration.

\textbf{2. Difficulty-Stratified Tasks:}
Tasks like \texttt{Car Painting}, \texttt{Star Battle}, and \texttt{Hitori} demonstrate clear separation between difficulty levels: easy instances are learned relatively early, while medium and hard variants require significantly more training to improve. This stratification indicates that the difficulty scaling mechanism is practical and yields meaningful distinctions in learning complexity.

\textbf{3. Stagnant or Low-Learning Tasks:}
Some tasks, including \texttt{Big Bench Symbolic} and \texttt{Magic Square}, show little to no improvement across all difficulty levels, particularly on the challenging setting. This suggests that these tasks may suffer from challenges such as sparse rewards, long-term dependencies, or overly complex solution spaces that PPO struggles to handle without additional guidance.

These learning patterns offer actionable insights for curriculum design. Tasks mastered early can serve as warm-up phases to bootstrap the agent's core skills, while more challenging tasks can be introduced later to push reasoning boundaries. Understanding which tasks exhibit positive transfer or potential interference is key to optimizing multi-task training regimes.

~\begin{table*}[hp]
    \centering
    \small
    \caption{Training data distribution across different strategies and stages.}
    \begin{tabular}{ccccc}
    \toprule
    \textbf{Method} & \textbf{\method-Data} & \textbf{ARC-AGI 1} & \textbf{ARC-AGI 2} & \textbf{AIME} \\
    \midrule
    Mix-Training & 11557 & 3160 & 5934 & 1738 \\
    Multi-Stage Stage 1 & 0 & 3160 & 5934 & 869 \\
    Multi-Stage Stage 2 & 11557 & 395 & 989 & 869 \\
    \bottomrule
    \end{tabular}
    \label{tab:030rldata}
\end{table*}
\section{Training Dataset Details}~\label{appendix:datadetails}
\subsection{Rejection Fine-tuning.}
For the puzzle part of the Rejection Finetuning (RFT) dataset, we sample 1,000 instances from each task in the \method dataset. 
We also include synthetic ARC-AGI data\footnote{https://github.com/neoneye/arc-dataset-collection/blob/main/dataset/ARC-Heavy/readme.md}.
We then use DeepSeek-R1 to generate 8 candidate solutions for each instance and select one correct solution per instance. 
The final puzzle dataset contains 12,041 high-quality puzzle samples. 
For the mathematical part of the RFT dataset, we collected mathematical problems from light-R1~\cite{wen2025light} that DeepSeek-R1 answered. We included all data from stage 2 of light-R1 and sampled 3,000 additional problems from stage 1, resulting in 12,533 mathematical samples.

\subsection{Reinforcement with Verifiable Puzzles.}
For our reinforcement learning with verifiable puzzles, we implemented two training paradigms. 
The detailed data mixing ratio is shown in Table~\ref{tab:030rldata} and introduced below.
As for the Mix-Training, the dataset for this approach consists of three components:
(1) \method-Train: 400 samples per task with equal distribution across difficulty levels. Note that we excluded eight tasks from the training set: six tasks deemed too difficult (Eight Puzzle, Fifteen Puzzle, Nine Puzzle, Sixteen Puzzle, Natural Language Navigation, Slant), one task too simple (Game24) based on pass@k evaluation, and one task lacking training data (FOLIO).
(2) ARC-AGI 1 and 2: Official datasets upsampled 8x to address specific reasoning challenges,
(3) AIME problems from 1983-2023 upsampled 2x as the mathematical component.
As for the Multi-stage RL, this approach follows a curriculum strategy with two distinct stages:
(1) Stage 1: Training on ARC-AGI 1 and 2 (upsampled 8x) along with AIME problems (upsampled 1x)
(2) Stage 2: Incorporating \method-Train while maintaining the datasets from Stage 1 (ARC-AGI 1, ARC-AGI 2, and AIME) without additional upsampling to prevent catastrophic forgetting

\section{Implementation Details}~\label{appendix:imdetails}

We finetune the model on this balanced dataset for 2 epochs using a maximum sequence length of 32768 tokens and a learning rate 1e-5, establishing a strong reasoning foundation before proceeding to reinforcement learning.

We adopt a variant of Proximal Policy Optimization (PPO)~\cite{schulman2017proximal}, i.e., VC-PPO~\cite{yuan2025s}, to train our reasoning agent on verifiable puzzles. This variant introduces several modifications tailored to long-chain-of-thought generation, improving training stability and performance over long sequences.

Standard PPO optimizes the following clipped objective:
\begin{equation}
\begin{aligned}
\mathcal{J}_\text{PPO}(\theta) = \mathbb{E}_{(q,a)\sim \mathcal{D},o_{\le t}\sim\pi_{\theta_{\text{old}}}} \Bigg[ 
\min \Bigg( r_t(\theta)\hat{A}_t, \ \text{clip}(r_t(\theta), 1 - \varepsilon, 1 + \varepsilon)\hat{A}_t \Bigg)
\Bigg],
\end{aligned}
\end{equation}
where $r_t(\theta) = \frac{\pi_{\theta}(o_t\mid q,o_{<t})}{\pi_{\theta_{\text{old}}}(o_t\mid q,o_{<t})}$ is the importance sampling ratio, and $\hat{A}_t$ is the estimated advantage. We compute $\hat{A}_t$ using Generalized Advantage Estimation (GAE)~\cite{schulman2017proximal}:
\begin{equation}
\hat{A}_t^{\text{GAE}(\gamma,\lambda)} = \sum_{l=0}^{\infty}(\gamma\lambda)^l\delta_{t+l},\quad \text{where} \quad
\delta_{l}=R_l+\gamma V(s_{l+1})-V(s_l).
\end{equation}

To optimize long-CoT training, we follow the VC-PPO and adopt the following modifications to the standard PPO algorithm:
\begin{itemize}
    \item \textbf{Removing KL Divergence Constraint.} In traditional RLHF~\cite{kaufmann2023survey}, a KL penalty is used to prevent the policy from diverging too far from the reference model. However, in long-CoT settings, this constraint often limits exploration and learning capacity. Following VC-PPO, we remove the KL term by setting $\texttt{kl\_loss\_weight} = 0.0$, allowing the model to deviate from the initial policy distribution freely.
    \item \textbf{Value Pretraining.} We observe that initializing the value model from the reward model causes unstable training due to their objective mismatch. To address this, we adopt value pretraining: first sample responses from a fixed SFT policy $\pi_{\text{sft}}$, compute Monte Carlo returns, and train the value model until the loss and explained variance converge. The pretrained value model is then used to initialize the critic for PPO.
    \item \textbf{Decoupled GAE.} We use different $\lambda$ values for the policy and the value model to reduce reward decay and improve optimization over long token sequences. Specifically, we set $\lambda_{\text{critic}} = 1.0$ for unbiased value estimation, and $\lambda_{\text{policy}} = 0.95$ to improve sample efficiency and learning speed.
\end{itemize}

\paragraph{Training Details.}
We set the maximum prompt length to 6,144 tokens and the maximum response length to 26,624 tokens. PPO training is conducted for 425 steps with a batch size 4,096 and a mini-batch size 512. The actor and critic are optimized using Adam, with learning rates of $1\times10^{-6}$ and $2\times10^{-6}$, respectively, and a linear warm-up schedule over 10 steps.
Before PPO begins, we perform value pretraining~\cite{yuan2025s} for 15 steps by collecting Monte Carlo returns from a fixed SFT policy and fitting the value model to these returns. 
Our implementation is based on the VeRL~\footnote{\url{https://github.com/volcengine/verl}.} framework.
We enable gradient checkpointing for both the actor and the critic to reduce memory consumption.
Rollouts are generated using temperature sampling ($\tau = 1.0$), with enforced end-of-sequence tokens. 
We leverage vLLM~\cite{kwon2023efficient} for efficient batched decoding with 256 rollout slots and paged attention.

\section{Baselines}\label{appendix:baselines}
We categorize our baseline models into the following groups for comparison:

\paragraph{General-purpose LLMs.} As reference benchmarks, we also included several top-tier general-purpose models:
\begin{itemize}
    \item GPT-4o-1120~\cite{gpt4}: OpenAI's model, representing the current highest standard of general models.
    \item Gemini 2.5 Pro~\cite{gemini}: Google's advanced model with excellent performance across various tasks.
    \item Claude 3.7 Sonnet~\cite{claude}: Anthropic's high-performance model known for its reliability and comprehensiveness.
    \item Grok-2-1212~\cite{grok}: xAI's open-weight model, demonstrating the potential of open-source models.
    \item Qwen-2.5-32B-Instruct~\cite{qwen2.5}: Alibaba's moderate-scale language model with strong performance in both Chinese and English tasks, which serves as our main backbone model for training.
\end{itemize}

\paragraph{Reasoning-specialized LLMs.} We selected the current state-of-the-art models specialized in reasoning:
\begin{itemize}
    \item o1~\cite{o1}, o3-mini~\cite{o1}, and o4-mini~\cite{o1}: Representing OpenAI's advancements in reasoning capabilities, particularly excelling in solving complex problems.
    \item DeepSeek-R1~\cite{guo2025deepseek}: A model optimized for mathematical and reasoning tasks, with outstanding performance in multi-step reasoning.
    \item Claude-3.7-Sonnet-Thinking~\cite{claude}: Employing a specialized chain-of-thought design to enhance capabilities in solving complex reasoning tasks.
    \item DeepSeek-R1-Distilled-Qwen2.5-32B~\cite{guo2025deepseek}: Transferring DeepSeek-R1's reasoning capabilities to a smaller model through knowledge distillation.
    \item QwQ-32B~\cite{qwq32b}: Alibaba's reasoning model trained from Qwen-2.5-32B.
\end{itemize}
This categorized comparison allows us to comprehensively evaluate our model's performance in both specialized reasoning capabilities and general abilities, while comparing against different types of state-of-the-art models.
For evaluation, we use temperature sampling ($\tau=1.0$).

\section*{Limitations}
\label{appendix:limitation}
Due to the limited time and resources, we did not train on other backbone models with \method-Data, or with RL algorithms other than VC-PPO.
However, we believe the results from our experiments can be generalized to other back models and algorithms.
Additionally, our data is all in single-turn textual form, and we did not include multi-turn puzzles or visual puzzles into \method we leave for future research.
Due to the research purpose of this work, we only train LLMs with hundreds of competitive math problems along with \method-Data.
In practice, our training resources could easily be included in an already comprehensive training set while maintaining (if not surpassing) the original performance of tasks in the original dataset.

\clearpage
\section{\method Details}\label{appendix:data_details}
This section details the \method dataset, including task specifications, difficulty estimation methodology, and example cases.

\begin{table*}[h]
    \caption{Details of 36 tasks in \method.}
    \centering
    \renewcommand{\arraystretch}{1.1} 
    \setlength{\tabcolsep}{6pt} 
    \scalebox{0.72}{
    \begin{tabular}{p{3.1cm} c c p{3cm} p{8.4cm}}
    \toprule
    \textbf{Task}  &
    \textbf{Categories}  &
    \textbf{Source} &
    \textbf{Difficulty Vriables} &
    \textbf{Rules}  \\
    \midrule

    \texttt{Binario} & Grid &     Auto   &\makecell {\textit{grid size} $n$,\\ \textit{mask rate} $r$,\\ \textit{minimal filled cells} $f$} & Given a partially filled binary grid, determine its unique completion such that each row and column contains an equal number of 0s and 1s, no more than two identical digits are adjacent, and all rows and columns are pairwise distinct.\\
    
    \cmidrule(lr){1-5} 
    \texttt{Campsite} & Grid &     Auto   &\makecell {\textit{grid height} $h$,\\ \textit{grid width} $w$,\\ \textit{tent count} $c$} & Given a grid with designated tree cells and empty cells, place tents on empty cells such that each tent is orthogonally adjacent to exactly one tree, no two tents are adjacent (including diagonally), and the number of tents in each row and column matches the specified totals.\\

    \cmidrule(lr){1-5} 
    \texttt{Magic Square} & Grid &     Auto   &\makecell {\textit{grid size} $n$,\\ \textit{mask rate} $r$ }
    & Complete the partially filled N×N magic square by assigning distinct integers to the empty cells such that all rows, columns, and both main diagonals sum to magic number, while preserving the given entries and satisfying all structural constraints.\\

    \cmidrule(lr){1-5} 
    \texttt{Skyscraper} & Grid &     Auto   &\makecell {\textit{grid size} $n$}
    &Given an N×N grid, assign each cell a unique building height from 1 to N per row and column, such that the number of visible buildings from each edge matches the provided visibility constraints, with taller buildings obscuring shorter ones. Return a valid configuration or report infeasibility.\\

    \cmidrule(lr){1-5} 
    \texttt{Sum Skyscraper} & Grid &     Auto   &\makecell {\textit{grid size} $n$}
    &Given an N X N grid, fill it with numbers 1 to N without repetition in any row or column, such that from each edge, the sum of visible building heights—where taller buildings block shorter ones behind—is equal to the corresponding clue.\\
    
    \cmidrule(lr){1-5} 
    \texttt{Star Battle} & Grid &     Auto   &\makecell {\textit{grid size} $n$,\\ \textit{number of star} $s$ }
    &Given an N×N grid composed of empty cells and blocked cells, place exactly one star in each row and each column such that stars occupy only empty cells and no two stars are adjacent horizontally, vertically, or diagonally.\\

    \cmidrule(lr){1-5} 
    \texttt{Sudoku2} & Grid &     Auto   &\makecell {\textit{mask rate} $r$ }
    &Given a partially filled 4×4 Sudoku grid with digits 1–4, fill in the remaining cells so that each row, column, and 2×2 subgrid contains each digit exactly once.\\

    \cmidrule(lr){1-5} 
    \texttt{Sudoku} & Grid &     Auto   &\makecell {\textit{mask rate} $r$ }
    &Given a partially filled 9×9 Sudoku grid, complete it so that each row, column, and 3×3 subgrid contains the digits 1 through 9 exactly once.\\

    \cmidrule(lr){1-5}  
    \texttt{Full Crosswords} & Grid &     ~\cite{jeggers_full_crossword_puzzles}   &\makecell {\textit{grid size} $n$}
    & Given a fixed crossword grid with blank and blocked cells, fill all blank cells with letters to form valid English words that satisfy the provided across and down clues, ensuring consistency at intersections.\\

    \cmidrule(lr){1-5}  
    \texttt{Symbolic Hard} & Grid &     ~\cite{bao2023assessing,10.24963/ijcai.2024/693}   & \textit{passrate/pass@k} & Given a symbolic 2D grid of numbers, learn the transformation rule across rows and apply it to generate the consistent output pattern, preserving vertical segment structures and alternating 0s in specific column bands.\\

    \cmidrule(lr){1-5} 
    \texttt{Crypto KKA} & Crypto &     Auto   &  
    \makecell{\textit{plaintext length
    } $l$, \\
    \textit{encryption method 
    } $e$, \\
    \textit{keyword length
    } $k$, \\
    \textit{shift in Caesar
    } $s$, \\
    \textit{rails in Rail Fence
    } $r$, \\
    \textit{a/b range in Affine
    } $r$, \\
    \textit{matrix size in Hill
    } $m$}
    &Known Key Attack-style Decryption. Given an encrypted ciphertext and the specification of an encryption method (e.g., Caesar, Vigenère, Hill, etc.), recover the original plaintext without being given the encryption algorithm or decryption procedure. The task requires understanding of classical ciphers and applying the correct decryption logic based on the cipher name and parameter settings.\\

    \cmidrule(lr){1-5} 
    \texttt{Crypto KPA} & Crypto &     Auto   &\makecell {\textit{plaintext length
    } $l$, \\
    \textit{encryption method 
    } $e$, \\
    \textit{keyword length
    } $k$, \\
    \textit{hint range 
    } $h$, \\
    \textit{shift in Caesar
    } $s$, \\
    \textit{rails in Rail Fence
    } $r$, \\
    \textit{a/b range in Affine
    } $r$, \\
    \textit{matrix size in Hill
    } $m$}
    &Known Plaintext Attack-style Decryption. Given a ciphertext and a plaintext-ciphertext example pair only, reverse-engineer the encryption transformation—without being given the algorithmic rule or key—and apply the inferred pattern to decrypt the target ciphertext. The task involves cryptanalytic generalization: learning a transformation from a single annotated example, and applying it to unseen input. \\

    \bottomrule
    \end{tabular}
    }
    \label{tab:appendix_task_details_1}
\end{table*}

\begin{table*}[h]
    \caption{Details of 36 tasks in \method.}
    \centering
    \renewcommand{\arraystretch}{1.1} 
    \setlength{\tabcolsep}{6pt} 
    \scalebox{0.72}{
    \begin{tabular}{p{3.1cm} c c p{3cm} p{8.4cm}}
    \toprule
    \textbf{Task}  &
    \textbf{Categories}  &
    \textbf{Source} &
    \textbf{Difficulty Vriables} &
    \textbf{Rules}  \\
    \midrule

    \texttt{Twiddle} & Sequential &     Auto   &\makecell {
    \textit{grid size} $n$ ,\\
    \textit{number of rotations} $r$}
    & Given an NXN grid of distinct numbers from 1 to n², restore it to row-major order using a sequence of counterclockwise kxk subgrid rotations, each specified by its top-left coordinate.\\

    \cmidrule(lr){1-5} 
    \texttt{Car Painting} & Sequential &     Auto   &\makecell {
    \textit{number of cars} $c$ ,\\
    \textit{number of color type} $t$ ,\\
    \textit{shift range} $k$,\\
    \textit{skew range} $s$}
    &Given a fixed initial sequence of cars to be painted, each car may be moved up to K positions forward or backward; the objective is to reorder the cars, within this constraint, to minimize adjacent color transitions (color switches).\\
    
    \cmidrule(lr){1-5}  
    \texttt{Stack Permutation} & Sequential &     Auto   &\makecell {
    \textit{sequence length} $l$ }& Given an input sequence, determine whether a target output sequence can be produced using a stack with only push (in order) and pop operations, ensuring last-in-first-out (LIFO) behavior.\\

    \cmidrule(lr){1-5} 
    \texttt{Big Bench Symbolic} & Sequential &     ~\cite{bao2023assessing,10.24963/ijcai.2024/693}   &\textit{passrate/pass@k} & Given a sequence of input-output list pairs, identify and apply the underlying symbolic transformation function to a new input list to produce its corresponding output.\\
    
    \cmidrule(lr){1-5} 
    \texttt{Eight Puzzle} & Sequential &     Auto   &\makecell {
    \textit{number of inversion} $n$ }& Given a 3×3 grid representing an 8-puzzle state, output the shortest sequence of moves (Up, Down, Left, Right) to reach the goal configuration [[1, 2, 3], [4, 5, 6], [7, 8, 0]], or indicate if no solution exists.\\
    
    \cmidrule(lr){1-5} 
    \texttt{Fifteen Puzzle} & Sequential &     Auto   &\makecell {
    \textit{number of inversion} $n$ } & Given a 4×4 grid representing a 15-puzzle state, output the shortest sequence of moves (U, D, L, R) to reach the goal configuration [[1, 2, 3, 4], [5, 6, 7, 8], [9, 10, 11, 12], [13, 14, 15, 0]], or report if no solution exists.\\

    \cmidrule(lr){1-5}  
    \texttt{Nine Puzzle} & Sequential &     Auto   &\makecell {
    \textit{number of inversion} $n$ }& Given a 3×3 grid of numbers 1–9, the player can circularly shift any row or column by 1 or 2 positions. Determine a sequence of moves that results in the grid sorted in ascending order, or report that it is unsolvable.\\

    \cmidrule(lr){1-5}  
    \texttt{Sixteen Puzzle} & Sequential &     Auto   &\makecell {
    \textit{number of inversion} $n$ }&  Given a 4×4 grid containing tiles numbered 1–16, players may circularly shift any row or column by 1 to 3 positions to sort the grid into ascending order; determine a valid move sequence or prove it unsolvable.\\

    \cmidrule(lr){1-5}  
    \texttt{Hitori} & Search &     Auto   &\makecell {
    \textit{grid size} $n$ }&  Given an NxN grid of numbers, black out cells to ensure each row and column contains no duplicate numbers, no two blacked cells are orthogonally adjacent, and all remaining white cells form a single connected group.\\

    \cmidrule(lr){1-5}  
    \texttt{Kakurasu} & Search &     Auto   &\makecell {
    \textit{grid size} $n$ ,\\
    \textit{black rate} $r$ }& Given a grid with row and column sum constraints, select black cells such that the sum of their positions in each row and column equals the respective targets, where each cell’s value is its 1-based index.\\

    \cmidrule(lr){1-5}  
    \texttt{Light Up} & Search &     Auto   &\makecell {
    \textit{grid size} $n$ ,\\
    \textit{black cell ratio} $r_1$ ,\\
    \textit{numbered ratio} $r_2$}& Given a rectangular grid with black numbered and unnumbered cells, place bulbs on empty cells to illuminate all white cells without lighting another bulb, ensuring each numbered black cell has exactly that many adjacent bulbs.\\

    \cmidrule(lr){1-5}  
    \texttt{Minesweeper} & Search &     Auto   &\makecell {
    \textit{board size} $n$ ,\\
    \textit{mine density} $d$ ,\\
    \textit{initial reveal ratio} $r$}&  Given a partially revealed Minesweeper grid, identify all unrevealed cells that must contain mines, based solely on the numerical clues and adjacency constraints.\\

    \cmidrule(lr){1-5}  
    \texttt{Slant} & Search &     Auto   &\makecell {
    \textit{board row} $r$ ,\\
    \textit{board col} $d$,\\
    \textit{hint ratio} $h$}&   Given a grid with numeric constraints at intersections, assign diagonal slashes(two directions) to each cell such that intersection counts are satisfied and no loops are formed.\\

    \cmidrule(lr){1-5} 
    \texttt{Checkmate in One} & Search &     ~\cite{srivastava2023beyond}   &\textit{passrate/pass@k} &Given a legal board configuration, find a move that results in immediate checkmate—i.e., the opposing king is placed in check and has no legal way to escape.\\

    \cmidrule(lr){1-5}  
    \texttt{Tic Tac Toe} & Search &
        Auto   &\makecell{
    \textit{board size} $n$ ,\\
    \textit{comparative potential} $p$\\
    \textit{center control} $c$,\\
    \textit{fork score} $f$}
    &  Given a partially filled N×N Tic Tac Toe board and the active player, identify the optimal move that maximizes the player’s winning chances or prevents immediate loss, according to standard game rules.\\

    \bottomrule
    \end{tabular}
    }
    \label{tab:appendix_task_details_2}
\end{table*}

\begin{table*}[h]
    \caption{Details of 36 tasks in \method.}
    \centering
    \renewcommand{\arraystretch}{1.1} 
    \setlength{\tabcolsep}{6pt} 
    \scalebox{0.72}{
    \begin{tabular}{p{3.1cm} c c p{3cm} p{8.4cm}}
    \toprule
    \textbf{Task}  &
    \textbf{Categories}  &
    \textbf{Source} &
    \textbf{Difficulty Vriables} &
    \textbf{Rules}  \\
    \midrule

    \texttt{Hamiltonian Cycle} & Graph &
    Auto
    &\makecell{
    \textit{number of nodes} $n$ ,\\
    \textit{edge density} $d$}
    & Given an undirected graph, determine whether a Hamiltonian cycle exists; if so, output one such cycle.\\

    \cmidrule(lr){1-5}  
    \texttt{Hamiltonian Path} & Graph &
    Auto 
    &\makecell{
    \textit{number of nodes} $n$ ,\\
    \textit{edge density} $d$}
    & Given an undirected graph, determine whether a Hamiltonian path exists; if so, output one such path.\\

    \cmidrule(lr){1-5}  
    \texttt{NL Navigation} & Graph &
    Auto 
    &\makecell{
    \textit{shortest path length} $l$ }& Given a spatial description of a road network among city landmarks, identify the shortest path from a designated starting point to the nearest landmark of a specified type.\\

    \cmidrule(lr){1-5}  
    \texttt{Maze} & Graph &
    Auto 
    &\makecell{
    \textit{obstacle percentage} $p$ }& Given a grid-based maze with designated start and end positions, the objective is to determine a valid path from start to end. Movement is restricted to the four cardinal directions (up, down, left, right), and traversal through blocked or impassable cells is not allowed.\\

    \cmidrule(lr){1-5}  
    \texttt{Knights and Knaves} &     Logic   &  ~\cite{szomiu2021puzzle} & 
    \makecell{\textit{ambiguity} $a$,\\ 
    \textit{number of inhabitants} $n$ }
    & Given statements from individuals who are either knights (truthful) or knaves (lying), decide if a specific conclusion is logically entailed, contradicted, or undetermined.\\

    \cmidrule(lr){1-5} 
    \texttt{FOLIO} & Logic &     ~\cite{han2022folio}   & 
    \makecell{\textit{number of primises} $n$ }
    & Given a set of premises, determine whether a conclusion logically follows from them.\\

    \cmidrule(lr){1-5} 
    \texttt{Zebra Logic} & Logic &     Auto   &\makecell {\textit{logic rule type} $t$,\\ 
    \textit{columns} $c$ ,\\
    \textit{rows} $r$ ,\\
    \textit{minimal conditions} $l$}
    &Given a fixed table structure and a set of categorical items with positional or equality constraints, deduce a unique one-to-one assignment of all elements that satisfies all logical conditions without transposing rows and columns.\\

    \cmidrule(lr){1-5}  
    \texttt{Game24} & Arithmetic &     Auto   &\makecell {
    \textit{number of integers} $n$ }& Given four to six integers, use each exactly once with +, –, ×, ÷, and parentheses to construct a valid expression that evaluates to 24.\\
    
    \cmidrule(lr){1-5}  
    \texttt{Countdown} & Arithmetic &     Auto   &\makecell {
    \textit{number of integers} $n$, \\
    \textit{range target value} $r$
    } & Given five integers and a target value, form a valid arithmetic expression using each number exactly once and the operators (+, –, ×, ÷), such that all intermediate results are positive integers and the final result equals the target.\\

    \bottomrule
    \end{tabular}
    }
    \label{tab:appendix_task_details_3}
\end{table*}

\subsection{Task Details}
We present the detailed specifications of all tasks in \method in Tables~\ref{tab:appendix_task_details_1}, \ref{tab:appendix_task_details_2}, and \ref{tab:appendix_task_details_3}. These tables provide comprehensive information about each task, including:
(1) Task category: \method encompasses seven distinct reasoning categories.
(2) Data source: Whether the task is automatically generated or sourced from existing datasets.
(3) Difficulty control variables: The specific parameters used to adjust task difficulty.
(4) Rule descriptions: Concise explanations of each puzzle type's rules.
These details illustrate the diversity and controllability of the \method dataset, highlighting how each task contributes to different aspects of reasoning assessment.

\setlength{\tabcolsep}{4pt}
\begin{table*}[h]
    \caption{Pass@k scores of GPT-4o.}
    \centering
\small
\begin{tabular}{l @{\hskip 4pt} ccc | ccc | ccc}
\toprule
\multicolumn{1}{c}{\textbf{Task Names}} &
\multicolumn{3}{c}{\textbf{Easy}} &
\multicolumn{3}{c}{\textbf{Medium}} &
\multicolumn{3}{c}{\textbf{Hard}} \\
\cmidrule(lr){2-4} \cmidrule(lr){5-7} \cmidrule(lr){8-10}

 & 
  \multicolumn{1}{c}{\textbf{pass@1}} &
  \multicolumn{1}{c}{\textbf{pass@10}}  &
  \multicolumn{1}{c}{\textbf{pass@100}} &
  \multicolumn{1}{c}{\textbf{pass@1}} &
  \multicolumn{1}{c}{\textbf{pass@10}}  &
  \multicolumn{1}{c}{\textbf{pass@100}} &
  \multicolumn{1}{c}{\textbf{pass@1}} &
  \multicolumn{1}{c}{\textbf{pass@10}}  &
  \multicolumn{1}{c}{\textbf{pass@100}} 
  \\
\midrule

Sudoku & {79.5} & {99.9} & {100.0} & {44.0} & {82.7} & {96.2} & {1.9} & {10.2} & {25.6} \\
Zebra Logic & {74.6} & {100.0} & {100.0} & {11.6} & {49.9} & {84.7} & {0.1} & {1.2} & {11.3} \\
Campsite & {25.5} & {87.3} & {100.0} & {4.7} & {33.0} & {75.0} & {0.7} & {6.3} & {44.6} \\
Crypto KKA & {51.9} & {82.8} & {97.4} & {35.4} & {65.4} & {82.5} & {18.5} & {45.7} & {57.5} \\
Crypto KPA & {17.5} & {36.5} & {54.2} & {8.1} & {13.7} & {23.1} & {5.7} & {20.8} & {29.6} \\
Maze & {36.2} & {88.5} & {100.0} & {12.1} & {49.0} & {83.6} & {9.0} & {37.1} & {67.2} \\
Magic Square & {85.0} & {100.0} & {100.0} & {46.3} & {88.3} & {99.9} & {1.6} & {12.9} & {38.4} \\

\bottomrule
\end{tabular}

    \label{tab:passk}
\end{table*}

\subsection{Difficulty Estimation}
In Section $\mathsection$~\ref{sec:data_construction}, we introduce our method for difficulty control in \method: 
determining different difficulty levels (easy, medium, hard) for puzzles through model pass@k metrics. 
Specifically, we use GPT-4o's pass@k (k=1,10,100) to establish these difficulty tiers. 
Table~\ref{tab:passk} showcases specific examples of puzzles at different difficulty levels. 
As demonstrated, there are clear distinctions in model pass@k performance between easy, medium, and hard levels across various tasks.

\clearpage
\subsection{Task Cases}
The following listings present examples of each puzzle type in the \method dataset. 
Each task includes a rule description and few-shot examples demonstrating puzzle mechanics. 
Color coding in prompts indicates different puzzle categories.
These examples highlight the diversity of our dataset, from grid-based challenges (Sudoku, Star Battle) to sequential puzzles (Eight Puzzle, Fifteen Puzzle). 
Each puzzle challenges distinct cognitive faculties, forming a comprehensive dataset spanning a wide spectrum of reasoning capabilities.

\lstset{
    backgroundcolor=\color[RGB]{245,245,244},
    breaklines=true,
    xleftmargin=5pt, 
    xrightmargin=5pt, 
    breakindent=0pt,
    basicstyle=\ttfamily\scriptsize,
    frame=trbl,
    frameround = tttt,
    emph={Rules, Task, Output, Format, Puzzle},
    emphstyle={\bfseries\color{Purple}}
}
\begin{lstlisting}[caption={Case of \texttt{Binario}},label=listing:case_binario]
You are tasked with solving a Binario puzzle.

Rules:
1. The Binario puzzle is played on a grid of size NxN, where N is an even number.
2. Each cell in the grid must be filled with either a 0 or a 1.
3. No more than half of the cells in any row or column can contain the same number.
4. No more than two identical numbers can be adjacent horizontally or vertically.
5. The puzzle must have a unique solution.

Task:
Solve the following Binario puzzle by filling in the missing cells (denoted by "_") with 0s and 1s according to the rules above.

Output Format:
Please output your answer within a code block (```) and format the grid as numbers, for example:

```
1 0 0 1  
0 1 1 0  
1 0 1 0  
0 1 0 1 
```

If no solution exists, output within the code block:

```
No valid solution exists for the given Binario puzzle.
```

Puzzle:
```
0 0 1 _
0 0 _ _
1 1 0 0
1 1 0 0
```

\end{lstlisting}

\lstset{
    backgroundcolor=\color[RGB]{245,245,244},
    breaklines=true,
    xleftmargin=5pt, 
    xrightmargin=5pt, 
    breakindent=0pt,
    basicstyle=\ttfamily\scriptsize,
    frame=trbl,
    frameround = tttt,
    emph={Rules, Task, Output, Format, Puzzle},
    emphstyle={\bfseries\color{Purple}}
}
\begin{lstlisting}[caption={Case of \texttt{Campsite}},label=listing:case_campsite]
You are tasked with solving a campsite puzzle.

Rules:
1. Notations
    (1) Trees are represented by `X`, tents are represented by `*`, and empty spaces are represented by `.`.
    (2) You will be given a board with trees and empty spaces, the total number of tents, and indications for the number of tents in each row and column. Your goal is to place tents on the empty spaces.
    
2. Constraints
    (1) Every tree on the board is associated with one tent, which is always horizontally or vertically adjacent to it.
    (2) No tent can be horizontally, vertically or diagonally adjacent to another tent.
    (3) The number of tents in each row and column should match the given indications.
    (4) The number of tents on the board should match the given total number of tents.
    (5) A tent can only be associated with one tree, but it can be adjacent to more than one tree.

Output Format:
1. Your output should include a solution followed by the final board.
2. You must not change trees (`X`) on the board, but only place tents (`*`) on empty spaces (`.`).
3. The final board should be wrapped between `<begin_board>` and `<end_board>` tags.

Task:
- Place tents on the empty spaces according to the given grid and rules.

Final Board:
```
<begin_board>
[Final Board]
<end_board>
```

Puzzle:
Here is the puzzle:
total number of tents: 4  
tents in each row: 1 1 0 2  
tents in each column: 1 1 1 1  
```
. X . .  
. . . .  
X X . X  
. . . .
``` 
\end{lstlisting}

\lstset{
    backgroundcolor=\color[RGB]{245,245,244},
    breaklines=true,
    xleftmargin=5pt, 
    xrightmargin=5pt, 
    breakindent=0pt,
    basicstyle=\ttfamily\scriptsize,
    frame=trbl,
    frameround = tttt,
    emph={Rules, Task, Output, Format, Puzzle},
    emphstyle={\bfseries\color{Purple}}
}
\begin{lstlisting}[caption={Case of \texttt{Magic Square}},label=listing:case_magic_square]
You are provided with a 3x3 Magic Square puzzle. Some cells are filled with numbers, while blank cells are represented by dots. 
Your task is to find a valid solution for the puzzle based on the following rules.

Rules:
1. Magic square is a 3x3 partially filled matrix.
2. You need to fill in the blanks in the matrix so that the sum of the numbers in each row, each column, and the two diagonals is equal.
3. You can only fill the blanks with integers, the filled matrix only consists of integers.
4. The filled numbers should not duplicate the already filled numbers.
5. Make sure that the sum of the numbers in each row, each column, and the two diagonals is equal.

Task:
- Fill the blank cells according to the given numbers and rules.
- Find a valid magic square solution for the given puzzle.

Output Format:
Please output your answer within a code block (```), formatted as a grid of numbers, for example:
```
2 4 3 1
3 1 2 4
4 3 1 2
1 2 4 3
```

Puzzle:
```
0 35 10  
. 15 5  
20 . 30
```

\end{lstlisting}

\lstset{
    backgroundcolor=\color[RGB]{245,245,244},
    breaklines=true,
    xleftmargin=5pt, 
    xrightmargin=5pt, 
    breakindent=0pt,
    basicstyle=\ttfamily\scriptsize,
    frame=trbl,
    frameround = tttt,
    emph={Rules, Task, Output, Format, Puzzle},
    emphstyle={\bfseries\color{Purple}}
}
\begin{lstlisting}[caption={Case of \texttt{Skyscraper}},label=listing:case_skyscraper]
Skycraper is a logic puzzle game where the goal is to fill a grid matrix based on the given clues. Here are the basic rules:

Rules:
1. Game Board: It typically consists of an n x n grid matrix. Each cell represents a building, and the building height is represented by a number ranging from 1 to n, where n is the size of the matrix.
2. Building Heights: Each row and column must be filled with numbers that represent building heights. Each number can only appear once in a row or column, similar to Sudoku constraints.
3. Visibility Clues: The hint numbers outside the matrix indicate how many buildings can be seen from that direction. Taller buildings block the view of shorter buildings behind them. Thus, a hint number represents how many buildings are visible from one end of a row or column.
For example, if the clue for a column is "3", it means that from the top or bottom of that column, 3 buildings can be seen, and the building heights must increase, as shorter buildings will be blocked by taller ones.
4. Objective: Fill the entire matrix based on the clues, ensuring that the heights of the buildings are distinct in each row and column and follow the visibility clues at the edges.
Example:

    [2] [1] [2] [3]
    +---+---+---+---+
[2] |   |   |   |   | [2]
    +---+---+---+---+
[3] |   |   |   |   | [2]
    +---+---+---+---+
[1] |   |   |   |   | [2]
    +---+---+---+---+
[4] |   |   |   |   | [1]
    +---+---+---+---+
    [2] [3] [2] [1]

This is an example of a Skycraper puzzle:
- The numbers at the top and bottom of the columns indicate how many buildings can be seen from that direction. For instance, the clue at the top of the first column is "2", meaning that 2 buildings can be seen from the top, indicating that at least one building is hidden by a taller one.
- The left and right clues for the rows work similarly, indicating how many buildings are visible from that row's left or right side.

Task:
Now, given a Skycraper puzzle, your task is to reconstruct the height of the buildings in each cell. If there are multiple solutions, return any one. If no valid solution exists, state that no solution exists.

Output Format:
- Please output your answer within a code block (```), formatted as a grid of numbers, for example:
```
2 4 3 1
3 1 2 4
4 3 1 2
1 2 4 3
```

Puzzle:
The input clues are:
Top:    3 2 2 1  
Left:   4 2 3 1  
Right:  1 2 2 2  
Bottom: 1 3 2 2
\end{lstlisting}

\lstset{
    backgroundcolor=\color[RGB]{245,245,244},
    breaklines=true,
    xleftmargin=5pt, 
    xrightmargin=5pt, 
    breakindent=0pt,
    basicstyle=\ttfamily\scriptsize,
    frame=trbl,
    frameround = tttt,
    emph={Rules, Task, Output, Format, Puzzle},
    emphstyle={\bfseries\color{Purple}}
}
\begin{lstlisting}[caption={Case of \texttt{Sum Skyscraper}},label=listing:case_sum_skyscraper]
Sum Skycraper is a logic puzzle game where the goal is to fill a grid matrix based on given clues and the height and number restrictions of rows and columns. Here are the basic rules:

Rules:
1. Game Board: Typically, it is an n x n grid matrix. Each cell represents a building, with its height represented by a number ranging from 1 to n, where n is the length of the matrix side.
2. Building Heights: Each row and column must be filled with numbers representing the heights of the buildings. Each number can only appear once in a row or column, similar to Sudoku constraints.
3. Visibility Clues: The hint numbers outside the matrix tell you how many buildings can be seen from that direction. Taller buildings will block shorter buildings behind them. Therefore, a hint number indicates the total height of buildings visible from one end of the row or column.
   For example, if a column's hint is "11," it means that looking from the top or bottom of that column, the total height of the visible buildings is 11. Additionally, the building heights need to be in increasing order, meaning shorter buildings are blocked by taller ones in front.
4. Objective: Fill the entire matrix according to the clues, ensuring that the heights of buildings in each row and each column are different, and that they comply with the visibility clues on the sides.

Example:

    [7] [4] [5] [9]
    +---+---+---+---+
[7] |   |   |   |   | [6]
    +---+---+---+---+
[9] |   |   |   |   | [5]
    +---+---+---+---+
[4] |   |   |   |   | [7]
    +---+---+---+---+
[10]|   |   |   |   | [4]
    +---+---+---+---+
    [5] [9] [7] [4]

The above is an example of a Sum Skycraper:
- The numbers at the top and bottom of the columns indicate how many buildings can be seen from that direction. For example, the hint at the top of the first column is "7," meaning that the total height of the visible buildings from top to bottom is 7.
- The hints on the left and right of the rows are similar to those for the columns, indicating the total height of buildings visible from the left and right sides of that row. For example, if the hint is 10, it means the visible buildings could possibly be 1, 2, 3, 4.

Task:
Now, given a Sum Skycraper scenario, please restore the height of each building in the scenario. If there are multiple answers, output any one of them. If there is no valid solution, then respond with "no valid solution."

Output Format:
- The input data consists of four lines: the first line represents the height seen from top to bottom, the second line from left to right, the third line from right to left, and the fourth line from bottom to top.
- Using the previous example, the input data would be:
```
7 4 5 9  
7 9 4 10  
6 5 7 4  
5 9 7 4
```
- Please output your answer within a code block (```), formatted as a grid of numbers storing the heights of the buildings, for example:
```
2 4 3 1
3 1 2 4
4 3 1 2
1 2 4 3
```
- If no solution exists, the result should be: "No valid solution."

Puzzle:
```
7 9 4 5  
7 9 4 5  
5 4 7 9  
5 4 7 9
```
Please provide the answer according to the above requirements.
\end{lstlisting}

\lstset{
    backgroundcolor=\color[RGB]{245,245,244},
    breaklines=true,
    xleftmargin=5pt, 
    xrightmargin=5pt, 
    breakindent=0pt,
    basicstyle=\ttfamily\scriptsize,
    frame=trbl,
    frameround = tttt,
    emph={Rules, Task, Output, Format, Puzzle},
    emphstyle={\bfseries\color{Purple}}
}
\begin{lstlisting}[caption={Case of \texttt{Star Battle}},label=listing:case_star_battle]
You are tasked with solving a customized version of the star-battle puzzle. The star-battle puzzle is a logic puzzle that requires the placement of stars in a grid.
There is only one solution to this puzzle.
Your goal is to determine the location of each star, adhering to the following rules:



Rules:
1. NOTATIONS:
    - The initial board consists of empty cells and blocked cells.
    - Empty cells are denoted by '.', blocked cells are denoted by 'X', and stars are denoted by '*'.
2. STAR MUST BE PLACED IN EMPTY CELL:
    - Each star must be placed in an EMPTY cell.
    - Blocked cells cannot contain stars.
    - You can only change cells denoted by '.', and must not change cells denoted by 'X'.
3. EXACTLY 1 STAR IN EACH ROW AND COLUMN:
    - Each row and column must contain EXACTLY one star.
    - No two stars can be in the same row or column.
    - There shouldn't be rows or columns without stars.
4. NO ADJACENT STARS ROW-WISE, COLUMN-WISE, OR DIAGONALLY:
    No two stars can be adjacent to each other, even diagonally.
    - Row-wise adjacency: two stars are in the same row, and there is no empty cell between them.
    - Column-wise adjacency: two stars are in the same column, and there is no empty cell between them.
    - Diagonal adjacency: two stars are in the same diagonal, and there is no empty cell between them.
    Note that you are prone to make mistakes in diagonal adjacency, so be extra careful. Here are 2 examples of diagonal adjacency in a partial grid:
    ```
    . * . .
    . . * .
    ```
    ```
    . . * .
    . * . .
    ```
    In both cases, the two stars are diagonally adjacent. You should avoid this situation.
5. CHECK FOR CONSTRAINTS AND BACKTRACK:
    - In each step, you should check if it violates the constraints in 2., 3., and 4.
    - If you find inconsistencies, you should backtrack and try a different placement.
    - If you find that you can't place a star anywhere without violating the constraints, you should backtrack and try a different placement.
    - Since there is only one solution, there is great chance your first placement is incorrect, which means you might need to start over.

Task:
Determin the location of each star and place them into the grid, adhering to the rules.

Output Format:
- Your output should include the final board.
- The final board is a revised version of the initial board in a way that if you need to place a star in an empty cell, replace the '.' with a '*'.
- Don't change the blocked cells denoted by 'X'.
- Your final board should be wrapped between <begin_board> and <end_board> tags.
- Use the following structure:

Final Board:
    ```
    <begin_board>
    [Final Board]
    <end_board>
    ```

Puzzle:
The input clues are:
```
. . . X .  
. . X . .  
. . X . .  
. . . X .  
. . X . X  
```
Please provide the answer according to the above requirements.
\end{lstlisting}

\lstset{
    backgroundcolor=\color[RGB]{245,245,244},
    breaklines=true,
    xleftmargin=5pt, 
    xrightmargin=5pt, 
    breakindent=0pt,
    basicstyle=\ttfamily\scriptsize,
    frame=trbl,
    frameround = tttt,
    emph={Rules, Task, Output, Format, Puzzle},
    emphstyle={\bfseries\color{Purple}}
}
\begin{lstlisting}[caption={Case of \texttt{Sudoku2}},label=listing:case_sudoku2]
You are provided with a 4x4 Sudoku puzzle. Some cells are filled with numbers, while empty cells are represented by dots. 
Your task is to find a valid solution for the puzzle based on the following rules:

Rules:
1. Board Structure: The Sudoku board is a 4x4 grid, divided into 4 smaller 2x2 subgrids (regions).
2. Number Range: Each cell can only contain a number between 1 and 
3. Row Rule: Each row must contain the numbers 1 through 4, with no repeats.
4. Column Rule: Each column must contain the numbers 1 through 4, with no repeats.
5. Subgrid Rule: Each 2x2 subgrid must contain the numbers 1 through 4, with no repeats.

Task:
- Find a valid Sudoku solution for the given puzzle.
- If there are multiple solutions, provide one.

Output Format:
- Please output your answer within a code block (```) representing the solved Sudoku board, formatted as a grid of numbers, for example:
```
2 4 3 1
3 1 2 4
4 3 1 2
1 2 4 3
```

Puzzle:
```
4 2 . 3
3 . . 4
2 . 4 1
1 4 3 2
```
Please provide the answer according to the above requirements.
\end{lstlisting}

\lstset{
    backgroundcolor=\color[RGB]{245,245,244},
    breaklines=true,
    xleftmargin=5pt, 
    xrightmargin=5pt, 
    breakindent=0pt,
    basicstyle=\ttfamily\scriptsize,
    frame=trbl,
    frameround = tttt,
    emph={Rules, Task, Output, Format, Puzzle},
    emphstyle={\bfseries\color{Purple}}
}
\begin{lstlisting}[caption={Case of \texttt{Sudoku}},label=listing:case_sudoku]
You are provided with a 9x9 Sudoku puzzle. Some cells are filled with numbers, while empty cells are represented by dots. 
Your task is to find a valid solution for the puzzle based on the following rules:

Rules:
1. Board Structure: The Sudoku board is a 9x9 grid, divided into 9 smaller 3x3 subgrids (regions).
2. Number Range: Each cell can only contain a number between 1 and 9.
3. Row Rule: Each row must contain the numbers 1 through 9, with no repeats.
4. Column Rule: Each column must contain the numbers 1 through 9, with no repeats.
5. Subgrid Rule: Each 3x3 subgrid must contain the numbers 1 through 9, with no repeats.

Task:
- Find a valid Sudoku solution for the given puzzle.
- If there are multiple solutions, provide one.

Output Format:
- Please output your answer within a code block (```) representing the solved Sudoku board, formatted as a grid of numbers, for example:
```
1 2 3 4 5 6 7 8 9
2 3 4 5 6 7 8 9 1
3 4 5 6 7 8 9 1 2
4 5 6 7 8 9 1 2 3
5 6 7 8 9 1 2 3 4
6 7 8 9 1 2 3 4 5
7 8 9 1 2 3 4 5 6
8 9 1 2 3 4 5 6 7 
9 1 2 3 4 5 6 7 8
```

Puzzle:
```
3 1 2 6 . . 5 9 4  
. 8 6 4 5 9 3 1 2  
5 9 4 2 3 1 7 8 6  
1 2 5 3 8 6 9 4 7  
8 6 3 7 9 4 1 2 .  
9 4 7 5 1 2 8 6 3  
4 7 8 . . . 6 . .  
6 . 1 8 . . 2 5 9  
2 5 9 . 6 . 4 7 8
```
Please provide the answer according to the above requirements.
\end{lstlisting}

\lstset{
    backgroundcolor=\color[RGB]{245,245,244},
    breaklines=true,
    xleftmargin=5pt, 
    xrightmargin=5pt, 
    breakindent=0pt,
    basicstyle=\ttfamily\scriptsize,
    frame=trbl,
    frameround = tttt,
    emph={Rules, Task, Output, Format, Puzzle},
    emphstyle={\bfseries\color{Purple}}
}
\begin{lstlisting}[caption={Case of \texttt{Full Crosswords}},label=listing:case_full_crosswords]
Task:
Your task is to complete the crossword puzzle grid based on the given clues. Ensure that all words are meaningful and semantically valid. The grid consists of fillable blank spaces ('_') and blocked spaces ('*'). You must strictly follow the provided grid layout, with no modifications allowed.

Rules:
1. Completing the Grid:
   - Fill each blank space ('_') with a letter to form valid words according to the given clues.
   - Blocked spaces ('*') must remain unchanged and cannot contain any letters.
   - The number of rows and columns must match the provided grid exactly.
   - All blank spaces ('_') must be filled--no empty spaces are allowed.

2. Clue Mapping Logic:
   (1) Across Clues:
     - Rows without '*' characters represent across words. These words correspond to across clues in order from top to bottom.
     - Rows containing '*' do not correspond to any across word or clue.

   (2) Down Clues:
     - Columns without '*' characters represent down words. These words correspond to down clues in order from left to right.
     - Columns containing '*' do not correspond to any down word or clue.

3. Matching Letters at Intersections:
   - Letters at the intersections of across and down words must match, ensuring valid words are formed both horizontally and vertically.

Output Format:
Please output your answer within a code block (```) as follows:

across: ANSWER1, ANSWER2, ...., ANSWER_N
down: ANSWER1, ANSWER2,....,ANSWER_N

- "across" contains the list of across words, ordered from top to bottom.
- "down" contains the list of down words, ordered from left to right.

Puzzle:
Across clues:
1. "London farewell!" (2001)  
2. "Fossil mollusk" (1972)  
3. "Radial's counterpart" (2013) 

Down clues:
1. "Measure of electric charge" (1999)  
2. "Is bookends?" (2014)  
3. Law-abiding (2010)  

The grid is as follows:
```
_ _ _ _ _ _ _
* _ _ * _ _ *
_ _ _ _ _ _ _
* _ _ * _ _ *
_ _ _ _ _ _ _
* _ _ * _ _ *
_ _ _ _ _ _ _
```
\end{lstlisting}

\lstset{
    backgroundcolor=\color[RGB]{245,245,244},
    breaklines=true,
    xleftmargin=5pt, 
    xrightmargin=5pt, 
    breakindent=0pt,
    basicstyle=\ttfamily\scriptsize,
    frame=trbl,
    frameround = tttt,
    emph={Rules, Task, and, Output, Format, Puzzle},
    emphstyle={\bfseries\color{Purple}}
}
\begin{lstlisting}[caption={Case of \texttt{Symbolic Hard}},label=listing:case_symbolic_hard]
Task and Rules:
Figure out the pattern in the following examples and apply it to the test case. Your answer must follow the format of the examples. 

Output Format:
Please output your answer within a code block (```), formatted as a grid of numbers.

Puzzle:

# Examples

Example 1:
[[2, 2, 2, 0, 2, 2, 2, 0, 2, 2, 2], 
 [2, 0, 2, 0, 2, 2, 2, 0, 2, 0, 2], 
 [2, 2, 2, 0, 2, 2, 2, 0, 2, 2, 2], 
 [2, 0, 2, 0, 2, 2, 2, 0, 2, 0, 2], 
 [2, 2, 2, 0, 2, 2, 2, 0, 2, 2, 2], 
 [2, 2, 2, 0, 2, 0, 2, 0, 2, 0, 2], 
 [2, 2, 2, 0, 2, 2, 2, 0, 2, 2, 2]] 
-> 
[[2, 2, 2, 0, 2, 2, 2, 0, 2, 2, 2], 
 [2, 2, 2, 0, 2, 0, 2, 0, 2, 0, 2], 
 [2, 2, 2, 0, 2, 2, 2, 0, 2, 2, 2], 
 [2, 2, 2, 0, 2, 0, 2, 0, 2, 0, 2], 
 [2, 2, 2, 0, 2, 2, 2, 0, 2, 2, 2], 
 [2, 0, 2, 0, 2, 2, 2, 0, 2, 0, 2], 
 [2, 2, 2, 0, 2, 2, 2, 0, 2, 2, 2]]

 Example 2:
[[2, 2, 2, 0, 2, 2, 2, 0, 2, 2, 2, 0, 2, 2, 2], 
 [2, 2, 2, 0, 2, 0, 2, 0, 2, 0, 2, 0, 2, 2, 2], 
 [2, 2, 2, 0, 2, 2, 2, 0, 2, 2, 2, 0, 2, 2, 2], 
 [2, 2, 2, 0, 2, 0, 2, 0, 2, 0, 2, 0, 2, 0, 2], 
 [2, 2, 2, 0, 2, 2, 2, 0, 2, 2, 2, 0, 2, 2, 2], 
 [2, 0, 2, 0, 2, 0, 2, 0, 2, 0, 2, 0, 2, 2, 2], 
 [2, 2, 2, 0, 2, 2, 2, 0, 2, 2, 2, 0, 2, 2, 2], 
 [2, 0, 2, 0, 2, 2, 2, 0, 2, 0, 2, 0, 2, 2, 2], 
 [2, 2, 2, 0, 2, 2, 2, 0, 2, 2, 2, 0, 2, 2, 2]] 
-> 
[[2, 2, 2, 0, 2, 2, 2, 0, 2, 2, 2, 0, 2, 2, 2], 
 [2, 2, 2, 0, 2, 2, 2, 0, 2, 0, 2, 0, 2, 0, 2], 
 [2, 2, 2, 0, 2, 2, 2, 0, 2, 2, 2, 0, 2, 2, 2], 
 [2, 0, 2, 0, 2, 2, 2, 0, 2, 0, 2, 0, 2, 0, 2], 
 [2, 2, 2, 0, 2, 2, 2, 0, 2, 2, 2, 0, 2, 2, 2], 
 [2, 2, 2, 0, 2, 0, 2, 0, 2, 0, 2, 0, 2, 0, 2], 
 [2, 2, 2, 0, 2, 2, 2, 0, 2, 2, 2, 0, 2, 2, 2], 
 [2, 2, 2, 0, 2, 0, 2, 0, 2, 2, 2, 0, 2, 0, 2], 
 [2, 2, 2, 0, 2, 2, 2, 0, 2, 2, 2, 0, 2, 2, 2]]

 Example 3:
[[2, 2, 2, 0, 2, 2, 2, 0, 2, 2, 2, 0, 2, 2, 2], 
 [2, 2, 2, 0, 2, 0, 2, 0, 2, 2, 2, 0, 2, 0, 2], 
 [2, 2, 2, 0, 2, 2, 2, 0, 2, 2, 2, 0, 2, 2, 2], 
 [2, 2, 2, 0, 2, 0, 2, 0, 2, 0, 2, 0, 2, 0, 2], 
 [2, 2, 2, 0, 2, 2, 2, 0, 2, 2, 2, 0, 2, 2, 2], 
 [2, 0, 2, 0, 2, 2, 2, 0, 2, 0, 2, 0, 2, 0, 2], 
 [2, 2, 2, 0, 2, 2, 2, 0, 2, 2, 2, 0, 2, 2, 2], 
 [2, 2, 2, 0, 2, 0, 2, 0, 2, 2, 2, 0, 2, 0, 2], 
 [2, 2, 2, 0, 2, 2, 2, 0, 2, 2, 2, 0, 2, 2, 2]] 
-> 
[[2, 2, 2, 0, 2, 2, 2, 0, 2, 2, 2, 0, 2, 2, 2], 
 [2, 2, 2, 0, 2, 2, 2, 0, 2, 0, 2, 0, 2, 0, 2], 
 [2, 2, 2, 0, 2, 2, 2, 0, 2, 2, 2, 0, 2, 2, 2], 
 [2, 2, 2, 0, 2, 0, 2, 0, 2, 0, 2, 0, 2, 0, 2], 
 [2, 2, 2, 0, 2, 2, 2, 0, 2, 2, 2, 0, 2, 2, 2], 
 [2, 0, 2, 0, 2, 0, 2, 0, 2, 2, 2, 0, 2, 0, 2], 
 [2, 2, 2, 0, 2, 2, 2, 0, 2, 2, 2, 0, 2, 2, 2], 
 [2, 2, 2, 0, 2, 2, 2, 0, 2, 0, 2, 0, 2, 0, 2], 
 [2, 2, 2, 0, 2, 2, 2, 0, 2, 2, 2, 0, 2, 2, 2]]

# Test Problem:
[[2, 2, 2, 0, 2, 2, 2, 0, 2, 2, 2, 0, 2, 2, 2, 0, 2, 2, 2], 
 [2, 2, 2, 0, 2, 2, 2, 0, 2, 0, 2, 0, 2, 0, 2, 0, 2, 0, 2], 
 [2, 2, 2, 0, 2, 2, 2, 0, 2, 2, 2, 0, 2, 2, 2, 0, 2, 2, 2], 
 [2, 2, 2, 0, 2, 2, 2, 0, 2, 2, 2, 0, 2, 0, 2, 0, 2, 2, 2], 
 [2, 2, 2, 0, 2, 2, 2, 0, 2, 2, 2, 0, 2, 2, 2, 0, 2, 2, 2], 
 [2, 0, 2, 0, 2, 2, 2, 0, 2, 0, 2, 0, 2, 0, 2, 0, 2, 0, 2], 
 [2, 2, 2, 0, 2, 2, 2, 0, 2, 2, 2, 0, 2, 2, 2, 0, 2, 2, 2], 
 [2, 2, 2, 0, 2, 0, 2, 0, 2, 0, 2, 0, 2, 0, 2, 0, 2, 0, 2], 
 [2, 2, 2, 0, 2, 2, 2, 0, 2, 2, 2, 0, 2, 2, 2, 0, 2, 2, 2], 
 [2, 0, 2, 0, 2, 2, 2, 0, 2, 2, 2, 0, 2, 0, 2, 0, 2, 0, 2], 
 [2, 2, 2, 0, 2, 2, 2, 0, 2, 2, 2, 0, 2, 2, 2, 0, 2, 2, 2]] 
-> 
[Your Answer Here]
\end{lstlisting}

\lstset{
    backgroundcolor=\color[RGB]{245,245,244},
    breaklines=true,
    xleftmargin=5pt, 
    xrightmargin=5pt, 
    breakindent=0pt,
    basicstyle=\ttfamily\scriptsize,
    frame=trbl,
    frameround = tttt,
    emph={Rules, Task, Output, Format, Puzzle},
    emphstyle={\bfseries\color{brown}}
}
\begin{lstlisting}[caption={Case of \texttt{Crypto KKA}},label=listing:case_crypto_kka]
Task:
Your task is to apply the provided decryption rules to decipher the given ciphertext.

Rules:
1. Review the decryption rules carefully to understand how the encryption method works.
2. Decrypt the provided ciphertext according to the rules, and derive the correct plaintext.

Output Format:
Please output your answer within a code block (```) as follows:

```
<result>
```
- <result> should be the decrypted plaintext corresponding to the given ciphertext, for example:
```
LOVE
```

Puzzle:
Ciphertext: DBUDIBUFOO
Encryption Method: {'method': 'Caesar Cipher', 'shift': 1}

What's the corresponding plaintext?
\end{lstlisting}

\lstset{
    backgroundcolor=\color[RGB]{245,245,244},
    breaklines=true,
    xleftmargin=5pt, 
    xrightmargin=5pt, 
    breakindent=0pt,
    basicstyle=\ttfamily\scriptsize,
    frame=trbl,
    frameround = tttt,
    emph={Rules, Task, Output, Format, Puzzle},
    emphstyle={\bfseries\color{brown}}
}
\begin{lstlisting}[caption={Case of \texttt{Crypto KPA}},label=listing:case_crypto_kpa]
Task:
Your task is to decrypt the given ciphertext and provide the corresponding plaintext. You will be given a sample hint that illustrates a pair of ciphertext and its matching plaintext to guide you in solving the puzzle.

Rules:
1. Analyze the provided ciphertext.
2. Use the sample hint as a reference to understand the encryption pattern or method used.
3. Apply the deciphering technique to convert the ciphertext into plaintext.

Output Format:
Please output your answer within a code block (```) as follows:

```
<result>
```
- <result> should be the decrypted plaintext corresponding to the given ciphertext, for example:
```
ABCIDEFG
```

Puzzle:
hint: Known plaintext and ciphertext pair:
Plaintext: 
```
AJORBRITISHSTUDYHASADDEDTOTHEEVIDENCETHATHORMONEREPLACEMENTTHERAPYINCREASESTHERISKOFBREASTC
ANCERESPECIALLYWHENWOMENRECEIVEACOMBINATIONOFESTRO
```
Ciphertext: 
```
NWBEOEVGVFUFGHQLUNFNQQRQGBGURRIVQRAPRGUNGUBEZBARERCYNPRZRAGGURENCLVAPERNFRFGUREVFXBSOERNFGP
NAPRERFCRPVNYYLJURAJBZRAERPRVIRNPBZOVANGVBABSRFGEB
```

Use the example above to decode:
```
RONYYGBERT
```
\end{lstlisting}

\lstset{
    backgroundcolor=\color[RGB]{245,245,244},
    breaklines=true,
    xleftmargin=5pt, 
    xrightmargin=5pt, 
    breakindent=0pt,
    basicstyle=\ttfamily\scriptsize,
    frame=trbl,
    frameround = tttt,
    emph={Rules, Task, Output, Format, Puzzle},
    emphstyle={\bfseries\color{blue}}
}
\begin{lstlisting}[caption={Case of \texttt{Twiddle}},label=listing:case_twiddle]
Given a 3x3 sliding puzzle where each cell contains a number (1 to 9), your goal is to restore the puzzle to its original sorted order through a series of rotation operations.

Rules:
1. You can select a 2x2 region within the 3x3 puzzle and rotate the positions of these 4 cells counterclockwise.
2. The goal is to restore the puzzle to its initial state (as shown below):

1 2 3  
4 5 6  
7 8 9

Task:
Please provide the steps to restore the puzzle to its initial state.

Output Format:
- Please output your answer within a code block (```) as follows:

```
<result>
```
- Replace `<result>` with a sequence of rotation steps, where each step is represented by a 2D coordinate (i, j) indicating the selection of the 2x2 region with (i, j) as the top-left corner for a counterclockwise rotation. 
For example:
```
(0,0)->(1,1)->(0,1)
```
- The problem guarantees that a solution exists.


Puzzle:
```
3 6 1
8 4 5
7 9 2
```
Please provide the answer according to the above requirements.
\end{lstlisting}

\lstset{
    backgroundcolor=\color[RGB]{245,245,244},
    breaklines=true,
    xleftmargin=5pt, 
    xrightmargin=5pt, 
    breakindent=0pt,
    basicstyle=\ttfamily\scriptsize,
    frame=trbl,
    frameround = tttt,
    emph={Rules, Task, Output, Format, Puzzle},
    emphstyle={\bfseries\color{blue}}
}
\begin{lstlisting}[caption={Case of \texttt{Car Painting}},label=listing:case_car_painting]
You are tasked with solving a Car Painting Problem.
In the Car Painting Problem, you play the role of a scheduler at a car painting factory. Your job is to arrange the painting sequence for a batch of cars to minimize the number of color switches, reducing paint waste and production time.

Rules:
1. There are N cars numbered from 1 to N that need to be painted.
2. Each car has a predetermined color (labeled as A, B, C, etc., for a total of M colors).
3. Cars enter the painting workshop in a fixed order, but can be rearranged within a range.
4. Each car can be moved forward or backward by at most K positions from its original position.
5. A color switch occurs when two adjacent cars have different colors, adding to the cost.
6. Your goal is to minimize the number of color switches by optimally arranging the cars.

Task:
Find a rearranged sequence of cars that minimizes the number of color switches.

You must provide a list of car IDs in their new order (rearranged to minimize color switches).

Puzzle:
Given the following information:
- Number of Cars (N): 14
- Number of Colors (M): 2
- Maximum Movement Range (K): 4
- Initial car sequence: [1, 2, 3, 4, 5, 6, 7, 8, 9, 10, 11, 12, 13, 14]
- Corresponding color: ['B', 'B', 'B', 'B', 'B', 'B', 'B', 'B', 'B', 'B', 'B', 'B', 'B', 'A']

Please find a rearranged car sequence that minimizes the number of color switches.
Remember: Each car can only be moved at most 4 positions forward or backward from its original position.

Output Format:
- Please output your answer within a code block (```), formatted as an array of integers representing the new order of car IDs, for example:

```
[2, 3, 1, 5, 8, 4, 6, 9, 7, 10]
```
- Replace `<result>` with a sequence of rotation steps, where each step is represented by a 2D coordinate (i, j) indicating the selection of the 2x2 region with (i, j) as the top-left corner for a counterclockwise rotation. 
For example:
```
(0,0)->(1,1)->(0,1)
```
- The problem guarantees that a solution exists.
\end{lstlisting}

\lstset{
    backgroundcolor=\color[RGB]{245,245,244},
    breaklines=true,
    xleftmargin=5pt, 
    xrightmargin=5pt, 
    breakindent=0pt,
    basicstyle=\ttfamily\scriptsize,
    frame=trbl,
    frameround = tttt,
    emph={Rules, Task, Output, Format, Puzzle},
    emphstyle={\bfseries\color{blue}}
}
\begin{lstlisting}[caption={Case of \texttt{Stack Permutation}},label=listing:case_stack_permutation]
Stack permutation is a permutation problem related to stacks (a last-in, first-out data structure). Given an input sequence, the goal is to generate different output sequences using stack operations (push and pop). A sequence is called a valid stack permutation if it can be obtained through stack operations.

Task and Rules:
Given a sequence of natural numbers, such as [1, 2, 3, 4], we want to know if a particular output sequence can be obtained through stack operations. The stack operations include:

1. Push: Add numbers from the input sequence to the stack in order.
2. Pop: Remove elements from the top of the stack and add them to the output sequence.

Example:

Suppose the input sequence is [1, 2, 3]. Here are some possible valid stack permutations:
- [1, 2, 3]: Directly push all elements into the stack and then pop them in order.  
- [2, 1, 3]: Push 1 and 2 into the stack, pop 2, then pop 1, and finally push and pop 3.  
- [3, 2, 1]: Push all elements into the stack, then pop them in reverse order.

Output Format:
- Please output your answer within a code block (```), for 
```
["Push(1)", "Pop()", "Push(2)", "Push(3)", "Pop()", "Pop()"]
```
- If the output sequence is not a valid stack permutation of the input sequence, output within the code block:
```
"The output sequence is not a valid stack permutation."
```

Puzzle:
Input sequence: [5, 2, 1, 3, 4]  
Output sequence: [5, 3, 1, 4, 2]

Please provide the solution according to the requirements below.
\end{lstlisting}

\lstset{
    backgroundcolor=\color[RGB]{245,245,244},
    breaklines=true,
    xleftmargin=5pt, 
    xrightmargin=5pt, 
    breakindent=0pt,
    basicstyle=\ttfamily\scriptsize,
    frame=trbl,
    frameround = tttt,
    emph={Rules, Task, and, Output, Format, Puzzle},
    emphstyle={\bfseries\color{blue}}
}
\begin{lstlisting}[caption={Case of \texttt{Big Bench Symbolic}},label=listing:case_big_bench_symbolic]
Task and Rules:
Apply a function to the final input list to generate the output list. Use any preceding inputs and outputs as examples to find what is the function used. All example outputs have been generated using the same function.

Output Format:
- Please output your answer within a code block (```), formatted as a list of numbers, for example:
[0, 2, 3]

Puzzle:
# Examples

Example 1:  
[7, 7, 9, 21, 7, 4, 4, 91, 0] -> [7, 9, 21, 7, 4, 4, 91, 0]

Example 2:  
[7, 78, 78, 7] -> [78, 78, 7]

Example 3:  
[9, 7, 72, 44, 7, 0, 7, 44] -> [9, 72, 44, 7, 0, 7, 44]

Example 4:  
[7, 8, 7, 7] -> [8, 7, 7]

# Test Problem:
[5, 37, 97, 48, 7, 1] ->
\end{lstlisting}

\lstset{
    backgroundcolor=\color[RGB]{245,245,244},
    breaklines=true,
    xleftmargin=5pt, 
    xrightmargin=5pt, 
    breakindent=0pt,
    basicstyle=\ttfamily\scriptsize,
    frame=trbl,
    frameround = tttt,
    emph={Rules, Task, Output, Format, Puzzle},
    emphstyle={\bfseries\color{blue}}
}
\begin{lstlisting}[caption={Case of \texttt{Eight Puzzle}},label=listing:case_8_puzzle]
Task:
The Eight puzzle is a classic sliding puzzle game. It consists of a 3x3 grid containing 8 numbered tiles from 1 to 8 and one blank space (represented by 0). The player moves these tiles with the ultimate goal of arranging them in order from 1 to 8. Below are the detailed rules:

Rules:
1. Initial State:
    - The initial state of the puzzle is 8 numbered tiles randomly distributed in a 3x3 grid, with the blank space located anywhere.
    - The puzzle usually starts from a scrambled state.
2. Movement:
    - The player can move a tile adjacent to the blank space into the blank space.
    - Tiles can only move in the four directions: up (U), down (D), left (L), and right (R).
    - Only one tile can be moved at a time.
3. Goal:
    - The ultimate goal is to arrange the tiles in order from left to right, top to bottom, as follows:
    
    1  2  3  
    4  5  6  
    7  8  0

Output Format:
- Please output your answer within a code block (```) as follows:
    ```
    <result>
    ```
    
- If there is an answer, the  is the sequence of moves, for example:
    ```
    LRURDL
    ```
- If there is no answer, the  is:
    ```
    No feasible move path exists.
    ```

Puzzle:
```
4  1  5
2  6  8
3  7  0
```
\end{lstlisting}

\lstset{
    backgroundcolor=\color[RGB]{245,245,244},
    breaklines=true,
    xleftmargin=5pt, 
    xrightmargin=5pt, 
    breakindent=0pt,
    basicstyle=\ttfamily\scriptsize,
    frame=trbl,
    frameround = tttt,
    emph={Rules, Task, Output, Format, Puzzle},
    emphstyle={\bfseries\color{blue}}
}
\begin{lstlisting}[caption={Case of \texttt{Fifteen Puzzle}},label=listing:case_15_puzzle]
Task:
The Fifteen puzzle is a classic sliding puzzle game. It consists of a 4*4 grid containing 15 numbered tiles from 1 to 15 and one blank space (represented by 0). The player moves these tiles with the ultimate goal of arranging them in order from 1 to 15. Below are the detailed rules:

Rules:
1. Initial State:
    - The initial state of the puzzle is 15 numbered tiles randomly distributed in a 4*4 grid, with the blank space located anywhere.
    - The puzzle usually starts from a scrambled state.
2. Movement:
    - The player can move a tile adjacent to the blank space into the blank space.
    - Tiles can only move in the four directions: up (U), down (D), left (L), and right (R).
    - Only one tile can be moved at a time.
3. Goal:
    - The ultimate goal is to arrange the tiles in order from left to right, top to bottom, as follows:
    
    1  2  3  4  
    5  6  7  8  
    9  10 11 12  
    13 14 15 0

Output Format:
- Please output your answer within a code block (```) as follows:
    ```
    <result>
    ```
    
- If there is an answer, the  is the sequence of moves, for example:
    ```
    LRURDL
    ```
- If there is no answer, the  is:
    ```
    No feasible move path exists.
    ```

Puzzle:
```
4   9   2   1
12  3   11  5
7   8   14  0
13  10  6  15
```
\end{lstlisting}

\lstset{
    backgroundcolor=\color[RGB]{245,245,244},
    breaklines=true,
    xleftmargin=5pt, 
    xrightmargin=5pt, 
    breakindent=0pt,
    basicstyle=\ttfamily\scriptsize,
    frame=trbl,
    frameround = tttt,
    emph={Rules, Task, Output, Format, Puzzle},
    emphstyle={\bfseries\color{blue}}
}
\begin{lstlisting}[caption={Case of \texttt{Nine Puzzle}},label=listing:case_nine_puzzle]
Task:
The nine puzzle is a classic sliding number puzzle. It consists of a 3x3 grid containing 9 tiles numbered from 1 to 9. Players can choose to move an entire row or column in a circular fashion each time. The ultimate goal is to arrange the tiles in numerical order from 1 to 9. The detailed rules are as follows:

Rules:
1. Initial State:
    - The initial state of puzzle consists of 9 number tiles randomly arranged on a 3x3 grid
    - The puzzle typically starts from a scrambled state.
2. Movement:
    - Players can choose to move an entire row or column, shifting it by 1 to 2 steps in a circular manner. For example: 1 2 3, shifting by 1 step results in 2 3 1, shifting by 2 steps results in 3 1 2.
    - We represent row moves as RAB, where A is the row number and B is the number of steps. Similarly, column moves are represented as CAB, where A is the column number and B is the number of steps.
3. Goal:
    - The ultimate goal is to arrange the tiles in order from left to right, top to bottom as follows:
    1  2  3  
    4  5  6  
    7  8  9

Output Format:
- If a solution exists, output the sequence of moves within a code block (```), for example:
    ```
    ["R11", "C23", "R32", "C12", "R23", "C31"]
    ```
- If no solution exists, output within the code block:
    ```
    "No valid sequence of moves exists."
    ```

Puzzle:
```
3  1  4  
5  7  6  
2  8  9
```
\end{lstlisting}

\lstset{
    backgroundcolor=\color[RGB]{245,245,244},
    breaklines=true,
    xleftmargin=5pt, 
    xrightmargin=5pt, 
    breakindent=0pt,
    basicstyle=\ttfamily\scriptsize,
    frame=trbl,
    frameround = tttt,
    emph={Rules, Task, Output, Format, Puzzle},
    emphstyle={\bfseries\color{blue}}
}
\begin{lstlisting}[caption={Case of \texttt{Sixteen Puzzle}},label=listing:case_sixteen_puzzle]
Task:
The sixteen puzzle is a classic sliding number puzzle. It consists of a 4x4 grid containing 16 tiles numbered from 1 to 16. Players can choose to move an entire row or column in a circular fashion each time. The ultimate goal is to arrange the tiles in numerical order from 1 to 16. The detailed rules are as follows:

Rules:
1. Initial State:
    - The initial state of the puzzle consists of 16 number tiles randomly arranged on a 4x4 grid.
    - The puzzle typically starts from a scrambled state.
2. Movement:
    - Players can choose to move an entire row or column, shifting it by 1 to 3 steps in a circular manner. For example: 1 2 3 4, shifting by 1 step results in 2 3 4 1, shifting by 2 steps results in 3 4 1 2, and shifting by 3 steps results in 4 1 2 3. 
    - We represent row moves as RAB, where A is the row number and B is the number of steps. Similarly, column moves are represented as CAB, where A is the column number and B is the number of steps.
3. Goal:
    - The ultimate goal is to arrange the tiles in order from left to right, top to bottom as follows:
    1  2  3  4  
    5  6  7  8  
    9  10 11 12  
    13 14 15 16

Output Format:
- If a solution exists, output the sequence of moves within a code block (```), for example:
    ```
    ["R11", "C23", "R32", "C12", "R23", "C31"]
    ```
- If no solution exists, output within the code block:
    ```
    "No valid sequence of moves exists."
    ```

Puzzle:
```
11  6  2  7  
12  4  5 16  
1   3 10 15  
8   9 13 14
```
\end{lstlisting}

\lstset{
    backgroundcolor=\color[RGB]{245,245,244},
    breaklines=true,
    xleftmargin=5pt, 
    xrightmargin=5pt, 
    breakindent=0pt,
    basicstyle=\ttfamily\scriptsize,
    frame=trbl,
    frameround = tttt,
    emph={Rules, Task, Output, Format, Puzzle},
    emphstyle={\bfseries\color{teal}}
}
\begin{lstlisting}[caption={Case of \texttt{Hitori}},label=listing:case_hitori]
You are tasked with solving a Hitori puzzle.

Rules:
1. The puzzle is played on an NxN grid (where N is an even number), with each cell containing a number.
2. Your goal is to "black out" certain cells, following these rules:
   - In each row and column, the same number cannot appear more than once. To eliminate repetitions, you must black out some of the cells.
   - Black cells cannot be adjacent, either horizontally or vertically.
   - All white cells (cells that are not blacked out) must be connected, meaning you can travel between any two white cells through horizontal or vertical moves.

Task:
- Solve the following Hitori puzzle by blacking out the cells where needed.
- You must provide the coordinates of the blacked-out cells.

Output Format:
- Please output your answer within a code block (```), formatted as a list of coordinates, for example:
    ```
    [(0, 0), (1, 3), (3, 2)]
    ```
- If no solution exists, output within the code block:
    ```
    "No valid solution exists for the given Hitori puzzle."
    ```

Puzzle:
```
[[5, 3, 1, 3, 5],  
 [1, 4, 3, 5, 3],  
 [1, 3, 2, 4, 3],  
 [3, 5, 1, 1, 2],  
 [3, 1, 4, 2, 2]]
```
\end{lstlisting}

\lstset{
    backgroundcolor=\color[RGB]{245,245,244},
    breaklines=true,
    xleftmargin=5pt, 
    xrightmargin=5pt, 
    breakindent=0pt,
    basicstyle=\ttfamily\scriptsize,
    frame=trbl,
    frameround = tttt,
    emph={Rules, Task, Output, Format, Puzzle},
    emphstyle={\bfseries\color{teal}}
}
\begin{lstlisting}[caption={Case of \texttt{Kakurasu}},label=listing:case_kakurasu]
You are tasked with solving a Kakurasu puzzle.

Task & Rules:
1. The puzzle is played on a rectangular grid (with arbitrary row and column sizes).
2. Your goal is to "black out" certain cells, following these rules:
   - The black cells in each row must sum up to the target number for that row.
   - The black cells in each column must sum up to the target number for that column.
   - To calculate the row sum:
     - The value of the black cells in each row is determined by their position in the row. For example, the first black cell in a row has a value of 1, the second black cell has a value of 2, and so on.
   - To calculate the column sum:
     - The value of the black cells in each column is determined by their position in the column. For example, the first black cell in a column has a value of 1, the second black cell has a value of 2, and so on.
3. Coordinates are 1-based. For example, the first row is row 1, and the first column is column 1. 

Puzzle:
Solve the following Kakurasu puzzle by blacking out the cells where needed.  
Board size: 4 X 4  
Row sums: [0, 5, 10, 5]  
Column sums: [5, 7, 7, 5]
You must provide the coordinates of the blacked-out cells.

Output Format:
- Please output your answer within a code block (```), formatted as a list of coordinates, for example:
    ```
    [(1, 1), (2, 4), (4, 3)]
    ```
- If no solution exists, output within the code block:
    ```
    "No valid solution exists for the given Kakurasu puzzle."
    ```

Puzzle:
```
[[5, 3, 1, 3, 5],  
 [1, 4, 3, 5, 3],  
 [1, 3, 2, 4, 3],  
 [3, 5, 1, 1, 2],  
 [3, 1, 4, 2, 2]]
```
\end{lstlisting}

\lstset{
    backgroundcolor=\color[RGB]{245,245,244},
    breaklines=true,
    xleftmargin=5pt, 
    xrightmargin=5pt, 
    breakindent=0pt,
    basicstyle=\ttfamily\scriptsize,
    frame=trbl,
    frameround = tttt,
    emph={Rules, Task, Output, Format, Puzzle},
    emphstyle={\bfseries\color{teal}}
}
\begin{lstlisting}[caption={Case of \texttt{Light Up}},label=listing:case_light_up]
You need to solve a Light Up puzzle.

Task & Rules:
1. The puzzle is played on a rectangular grid (the number of rows and columns is not fixed).
2. The goal is to place light bulbs (represented by L) on the empty squares of the grid, following these rules:
   - Each numbered black square (represented by numbers 1-4) must have the specified number of light bulbs around it. For example, a black square with "1" means it must have exactly 1 light bulb around it, a "2" means exactly 2 light bulbs, and so on.
   - Light bulbs can only be placed on empty squares (represented by .) and must only light up empty squares; they cannot light up other light bulbs.
   - Black squares (represented by #) cannot be illuminated and block the light of the light bulbs.

Puzzle:
Solve the following Light Up puzzle by placing the required light bulbs (L):  
.#..#.  
...1..  
#.....  
...1..  
.....#  
......

You need to provide the coordinates of the light bulbs in the following format: 

Output Format:
- - Please output your answer within a code block (```), formatted as a list of numbers, for example:
    ```
    [(2, 4), (6, 2), (7, 5), (0, 0), (4, 3), (1, 1), (2, 0), (5, 1), (1, 7), (3, 2)]
    ```
- If no solution exists, output within the code block:
    ```
    "No solution, unable to solve the given Light Up puzzle."
    ```
\end{lstlisting}

\lstset{
    backgroundcolor=\color[RGB]{245,245,244},
    breaklines=true,
    xleftmargin=5pt, 
    xrightmargin=5pt, 
    breakindent=0pt,
    basicstyle=\ttfamily\scriptsize,
    frame=trbl,
    frameround = tttt,
    emph={Rules, Task, Output, Format, Puzzle},
    emphstyle={\bfseries\color{teal}}
}
\begin{lstlisting}[caption={Case of \texttt{Minesweeper}},label=listing:case_minesweeper]
You are playing a Minesweeper game.
On a grid composed of different cells, the player's goal is to deduce all cells that contain mines and avoid clicking on mines.
Each cell in the game may either contain a mine or show the number of adjacent mines.

Rules:
1. Grid and Mines
The game grid consists of several cells, each of which may be:
    - Mine: If the player clicks on a mine cell, the game ends.
    - Numbered cell: This cell displays the number of mines adjacent to it. The number indicates how many of the eight neighboring cells contain mines.
2. Current Grid State Representation
The grid state is represented as:
    -2: Indicates the cell is unknown (not revealed).
    0-8: Revealed non-mine cells, where the number indicates how many mines are adjacent to that cell. For example, a cell with 0 means no mines adjacent, a cell with 1 means one mine adjacent, and so on.

Task:
Based on current grid state, infer positions of the mines that can be definitively determined.

Puzzle:
The current grid state is:
[[-2, 1, -2, -2, -2, -2, -2, -2, 0], 
 [-2, -2, -2, 0, 0, -2, 0, -2, 0], 
 [-2, 2, -2, -2, -2, -2, -2, -2, -2], 
 [-2, 2, -2, -2, -2, 0, -2, -2, -2], 
 [-2, -2, -2, -2, 1, 1, -2, -2, -2], 
 [-2, -2, -2, -2, 1, -2, 1, -2, -2], 
 [1, -2, 1, -2, -2, -2, 2, 1, -2], 
 [2, -2, -2, -2, -2, 0, 1, -2, -2], 
 [-2, -2, -2, 1, -2, 0, 1, 1, 1]]

Coordinate Explanation:
- Coordinates start from (0, 0), where the first row, first column is (0, 0), the second row, second column is (1, 1), and so on.
- You need to provide the coordinates of the determinable mines in the following format:

Output Format:
- Please output your answer within a code block (```), formatted as a list of coordinates (r, c), for example:

    ```
    [(1, 1), (2, 4), (4, 3)]
    ```
    
- If no solution exists, output within the code block:

    ```
    "Unable to determine any mine locations."
    ```
\end{lstlisting}

\lstset{
    backgroundcolor=\color[RGB]{245,245,244},
    breaklines=true,
    xleftmargin=5pt, 
    xrightmargin=5pt, 
    breakindent=0pt,
    basicstyle=\ttfamily\scriptsize,
    frame=trbl,
    frameround = tttt,
    emph={Rules, Task, Output, Format, Puzzle},
    emphstyle={\bfseries\color{teal}}
}
\begin{lstlisting}[caption={Case of \texttt{Slant}},label=listing:case_guess]
You need to solve a Slant puzzle.

Task & Rules:
1. Grid Numbers:
   - Each cell in the grid may contain a number, indicating how many diagonal lines meet at that intersection. The number ranges from 0 to 4, representing the number of intersecting diagonal lines.
2. Diagonal Line Rules:
   - Each cell must contain one diagonal line, either a "/" (forward slash, representing top-left to bottom-right) or a "\" (backslash, representing top-right to bottom-left).
3. Intersection Numbers:
   - The number indicates how many diagonal lines meet at that intersection. For example:
     - Number 1: Indicates 1 diagonal line intersects at that point.
     - Number 2: Indicates 2 diagonal lines intersect at that point.
     - Number 0: Indicates no diagonal lines intersect at that point.
     - Numbers 3 and 4: Represent 3 and 4 intersecting diagonal lines, respectively.
4. No Loops:
   - The diagonal lines must not form loops. All diagonal lines must connect, and no closed cycle can be formed.

Puzzle: 
Solve the following slant puzzle:  
1 0 1 . 0 2 0  
1 3 1 1 4 0 2  
1 1 4 2 0 . 0  
1 . 2 2 3 1 1  
0 4 0 2 3 2 1  
2 1 3 2 0 3 1  
0 . . 1 2 0 1

Output Format:
- Please output your answer within a code block (```), formatted as a grid of numbers, for example:
    ```
    1  1  1 1
    -1 1 -1 1
    1 -1 -1 1
    ```
- 1 represents "/" (forward slash, top-left to bottom-right)  
- -1 represents "\" (backslash, top-right to bottom-left)
\end{lstlisting}

\lstset{
    backgroundcolor=\color[RGB]{245,245,244},
    breaklines=true,
    xleftmargin=5pt, 
    xrightmargin=5pt, 
    breakindent=0pt,
    basicstyle=\ttfamily\scriptsize,
    frame=trbl,
    frameround = tttt,
    emph={Rules, Task, Output, Format, Puzzle},
    emphstyle={\bfseries\color{teal}}
}
\begin{lstlisting}[caption={Case of \texttt{Checkmate in One}},label=listing:case_checkmate_in_one]
Task:
You are tasked with finding a move in the chess position resulting in checkmate:

Output Format:
- Please output your answer within a code block (```) as follows, for example:
```
Rg5#
```

Puzzle:
Here is the chess position:
1. c4 c5  
2. g3 e6  
3. Bg2 d5  
4. cxd5 exd5  
5. Nc3 Nf6  
6. Nf3 b6  
7. d4 c4  
8. O-O Bb7  
9. Ne5 Bd6  
10. Bf4 O-O  
11. Qc2 Nc6  
12. Nxd5 Nxd4  
13. Nxf6+ Kh8  
14. Qxc4 Bxe5  
15. Bxe5 Rc8  
16. Qxd4 Qe7  
17. Nd5 Bxd5  
18. Bxg7+ Kg8  
19. Qxd5 Rfd8  
20. Qe5 Qb4  
21. Rad1 Re8  
22. Qg5 Qe4  
23. Bh6+ Kh8  
24.
\end{lstlisting}

\lstset{
    backgroundcolor=\color[RGB]{245,245,244},
    breaklines=true,
    xleftmargin=5pt, 
    xrightmargin=5pt, 
    breakindent=0pt,
    basicstyle=\ttfamily\scriptsize,
    frame=trbl,
    frameround = tttt,
    emph={Rules, Task, Output, Format, Puzzle},
    emphstyle={\bfseries\color{teal}}
}
\begin{lstlisting}[caption={Case of \texttt{Tic Tac Toe}},label=listing:case_tic_tac_toe]
You are tasked with solving a Tic Tac Toe puzzle.

Task & Rules:
1. The board consists of 3x3 cells.
2. Players take turns placing their mark on an empty cell, one move per turn. The two players use "O" or "X".
3. A player wins by placing three of their marks consecutively in a row, column, or diagonal.
4. If the board is completely filled without a winner, the game is a draw.

You are playing tic-tac-toe as X. 

Puzzle:
Current board:  
O | X |   
---------  
  |   | O  
---------  
X | X | O

Question: What is the best next move? Please provide only your move and display the board.


Output Format:
- Please output your answer within a code block (```), for example:
```
"X" "O" ""
"" "X" ""
"O" "" "X"
\end{lstlisting}

\lstset{
    backgroundcolor=\color[RGB]{245,245,244},
    breaklines=true,
    xleftmargin=5pt, 
    xrightmargin=5pt, 
    breakindent=0pt,
    basicstyle=\ttfamily\scriptsize,
    frame=trbl,
    frameround = tttt,
    emph={Rules, Task, Output, Format, Puzzle},
    emphstyle={\bfseries\color{Fuchsia}}
}
\begin{lstlisting}[caption={Case of \texttt{Game24}},label=listing:case_game24]
You are given four or five or six integers, ranging from 1 to 13, provide an arithmetic expression that results in 24.

Task & Rules:
You must use all the given numbers, each exactly once.  
The operators you can use include: addition (+), subtraction (-), multiplication (*), and division (/).  
You can use parentheses to change the order of operations.

Output Format:
Please output your answer within a code block (```) as follows:
```
<result>
```
- If there is a solution, <result> is the sequence of numbers and operators that results in 24, for example: 
```
(8 / 2) * (8 - 2)
```
- If there is no solution, <result> is "cannot form 24".

Puzzle:
Input: 7, 1, 7, 13
Please provide a solution for the 24 game according to the above rules and input.
\end{lstlisting}

\lstset{
    backgroundcolor=\color[RGB]{245,245,244},
    breaklines=true,
    xleftmargin=5pt, 
    xrightmargin=5pt, 
    breakindent=0pt,
    basicstyle=\ttfamily\scriptsize,
    frame=trbl,
    frameround = tttt,
    emph={Rules, Task, Output, Format, Puzzle},
    emphstyle={\bfseries\color{Fuchsia}}
}
\begin{lstlisting}[caption={Case of \texttt{Countdown}},label=listing:case_countdown]
You are given 5 integers and a target number. Your task is to create an arithmetic expression that results in exactly the target number.

Task & Rules:
- You must use ALL the given numbers, each exactly once.
- The operators you can use include: addition (+), subtraction (-), multiplication (*), and division (/).
- You can use parentheses to change the order of operations.
- All intermediate results must be positive integers (no fractions or negative numbers allowed).

Output Format:
Please output your answer within a code block (```) as follows:
```
<result>
```
- If there is a solution, <result> is the sequence of numbers and operators that results in 24, for example: 
```
(8 / 2) * (8 - 2)
```
- If there is no solution, <result> is "cannot form 85".

Puzzle:
Numbers: 10, 5, 15, 2, 9
Target: 85
Please provide a solution for the 24 game according to the above rules and input.
\end{lstlisting}

\lstset{
    backgroundcolor=\color[RGB]{245,245,244},
    breaklines=true,
    xleftmargin=5pt, 
    xrightmargin=5pt, 
    breakindent=0pt,
    basicstyle=\ttfamily\scriptsize,
    frame=trbl,
    frameround = tttt,
    emph={Rules, Task, Output, Format, Puzzle},
    emphstyle={\bfseries\color{Green}}
}
\begin{lstlisting}[caption={Case of \texttt{Hamiltonian Cycle}},label=listing:case_hamiltonian_cycle]
You are tasked with solving a Hamiltonian Cycle Puzzle.

Task & Rules:
A Hamiltonian Cycle in an undirected graph is a cycle that visits every vertex exactly once and returns to the starting vertex. The task is to determine whether a Hamiltonian Cycle exists in the given graph.

The graph is represented as follows:
- The first line contains a single integer `N`, the number of vertices in the graph.
- The subsequent lines each describe an edge in the graph. Each edge is represented by two space-separated integers `u` and `v`, which indicate that there is an undirected edge between vertex `u` and vertex `v`.
- The vertices are numbered from `0` to `N-1`.

Output Format:
- Please output your answer within a code block (```) as follows:
```
<result>
```
```
- If a Hamiltonian Cycle exists, <result> should be a list of vertex indices that form the cycle, where the last vertex is the same as the first vertex to complete the cycle, for example:
```
[0, 2, 3, 1, 0]
```
- If no Hamiltonian Cycle exists, <result> should be "NO".

Puzzle:
11
0 1
0 4
2 3
0 2
6 10
1 10
1 3
\end{lstlisting}

\lstset{
    backgroundcolor=\color[RGB]{245,245,244},
    breaklines=true,
    xleftmargin=5pt, 
    xrightmargin=5pt, 
    breakindent=0pt,
    basicstyle=\ttfamily\scriptsize,
    frame=trbl,
    frameround = tttt,
    emph={Rules, Task, Output, Format, Puzzle},
    emphstyle={\bfseries\color{Green}}
}
\begin{lstlisting}[caption={Case of \texttt{Hamiltonian Path}},label=listing:case_hamiltonian_path]
You are tasked with solving a Hamiltonian Path Puzzle.

Task & Rules:
A Hamiltonian Path in an undirected graph is a path that visits every vertex exactly once. The task is to determine whether a Hamiltonian Path exists in the given graph.

The graph is represented as follows:
- The first line contains a single integer `N`, the number of vertices in the graph.
- The subsequent lines each describe an edge in the graph. Each edge is represented by two space-separated integers `u` and `v`, which indicate that there is an undirected edge between vertex `u` and vertex `v`.
- The vertices are numbered from `0` to `N-1`
    
Output Format:
- Please output your answer within a code block (```) as follows:
```
<result>
```
- <result> should be a list of vertex indices that form the Hamiltonian Path if it exists, for example:
```
[0, 2, 3, 1, 0]
```
- If no Hamiltonian Path exists, <result> should be "NO".

Puzzle:
14
0 4
0 5
0 8
0 9
1 3
1 6
1 9
2 5
2 9
2 12
2 13
3 6
3 9
3 10
4 11
5 12
5 13
6 10
6 11
6 13
7 8
7 11
7 13
8 10
9 12
11 12
\end{lstlisting}

\lstset{
    backgroundcolor=\color[RGB]{245,245,244},
    breaklines=true,
    xleftmargin=5pt, 
    xrightmargin=5pt, 
    breakindent=0pt,
    basicstyle=\ttfamily\scriptsize,
    frame=trbl,
    frameround = tttt,
    emph={Rules, Task, Output, Format, Puzzle},
    emphstyle={\bfseries\color{Green}}
}
\begin{lstlisting}[caption={Case of \texttt{NL Navigation}},label=listing:case_nl_navigation]
You are tasked with solving a spatial reasoning puzzle involving navigation in a city. Follow these guidelines to determine the shortest path to a specific landmark:

Task & Rules:
1. Landmarks Definition:
   - Identify a set of landmarks which include: store, bank, house, cinema, garden, and school.
   - The total number of landmarks in the puzzle will range from 7 to 10.
2. Structure:
   - The landmarks are organized in a binary tree structure.
   - The root node of this tree represents the starting point for navigation.
3. Objective:
   - Your goal is to find the shortest path from the starting point to the nearest specified type of landmark.
4. Puzzle Input:
   - You will receive a question.
   - Use the information provided in the question to determine the path.
    
Output Format:
- Please output your answer "([A-Z, ]+)" within a code block (```), containing only the path letters, for example:

```
EFJ
```

- Your answer should only include the uppercase letters representing the landmarks in the path.
- If the path is direct with no intermediate landmarks, provide an empty code block:
```
```

Puzzle:
Here is the puzzle:  
Story: There is a set of roads and a set of landmarks. The start point is cinema H.  
There is a road which is 100 meters long from cinema H to house F.  
There is a road which is 200 meters long from house F to store I.  
There is a road which is 200 meters long from store I to cinema A.  
There is a road which is 100 meters long from cinema A to bank J.  
There is a road which is 100 meters long from bank J to house B.  
There is a road which is 200 meters long from cinema H to house D.  
There is a road which is 100 meters long from house D to bank E.  
There is a road which is 200 meters long from house D to cinema C.

Question: From the start point, how to reach the nearest bank?

Please provide the solution according to the requirements above.
\end{lstlisting}

\lstset{
    backgroundcolor=\color[RGB]{245,245,244},
    breaklines=true,
    xleftmargin=5pt, 
    xrightmargin=5pt, 
    breakindent=0pt,
    basicstyle=\ttfamily\scriptsize,
    frame=trbl,
    frameround = tttt,
    emph={Rules, Task, Output, Format, Puzzle},
    emphstyle={\bfseries\color{Green}}
}
\begin{lstlisting}[caption={Case of \texttt{Maze}},label=listing:case_maze]
You are tasked with solving a Maze Puzzle.

Puzzle:
Given a 5x5 maze map, as shown below:

S . . B .
B . . . .
. . . . B
. . . . .
. . . . E

Where:
S represents the start point (located in the top-left corner at coordinates (1, 1))  
E represents the end point (located in the bottom-right corner at coordinates (5, 5))  
B represents an obstacle (impassable)  
. represents open space (passable)

Rules:
1. You can only move up, down, left, or right, not diagonally.  
2. You cannot pass through obstacles (B).  
3. You can move freely on open spaces (.).  
4. The goal is to find a path from the start point (S) to the end point (E).

Please find a valid path from the start point (S) to the end point (E). If there are multiple paths, provide any one of them. If no valid path exists, state that it is impossible to reach the end point.

Output Format:
- Please output your answer within a code block (```) as follows:
```
<result>
```
- If there is a path, <result> is the sequence of coordinates in the path. For example: 
```
(1,1)->(1,3)->(3,5)
```
- If no path exists, output directly: 
```
not exist the path from start to end.
```
Please provide the solution according to the requirements above.
\end{lstlisting}

\lstset{
    backgroundcolor=\color[RGB]{245,245,244},
    breaklines=true,
    xleftmargin=5pt, 
    xrightmargin=5pt, 
    breakindent=0pt,
    basicstyle=\ttfamily\scriptsize,
    frame=trbl,
    frameround = tttt,
    emph={Rules, Task, Output, Format, Puzzle},
    emphstyle={\bfseries\color{Goldenrod}}
}
\begin{lstlisting}[caption={Case of \texttt{Knights and Knaves}},label=listing:case_knights_and_knaves]
In this puzzle, you are presented with a scenario involving inhabitants of an island where each person is either a knight or a knave. Knights always tell the truth, while knaves always lie. Your task is to determine the truth value of a given statement based on the information provided.

Task & Rules:
1. Knights always tell the truth.
2. Knaves always lie.
3. Use logical reasoning to determine the truth value of the statement.

Output Format:
- Please output your answer within a code block (```) as follows:
```
<result>
```
Options:
- "Entailment": Use this if the statement is logically true based on the information provided.
- "Contradiction": Use this if the statement contradicts known facts or logical deductions.
- "Unknown": Use this if the truth value of the statement cannot be determined with the given information.

Puzzle:
On the island where each inhabitant is either a knave or a knight, knights always tell the truth while knaves always lie.
You meet four inhabitants: Alice, Bill, Ted, and Mel.
    - Bill tells you that Mel and Ted are not the same.
    - Mel claims that it is false that Alice is a knave.
    - Ted says that Alice is a knight and Mel is a knave.
    - Alice tells you that only a knave would say that Ted is a knave.

Can you determine who is a knight and who is a knave?
Question: Is Ted the knight?
\end{lstlisting}

\lstset{
    backgroundcolor=\color[RGB]{245,245,244},
    breaklines=true,
    xleftmargin=5pt, 
    xrightmargin=5pt, 
    breakindent=0pt,
    basicstyle=\ttfamily\scriptsize,
    frame=trbl,
    frameround = tttt,
    emph={Rules, Task, Output, Format, Puzzle},
    emphstyle={\bfseries\color{Goldenrod}}
}
\begin{lstlisting}[caption={Case of \texttt{FOLIO}},label=listing:case_folio]
Analyze the given premises and determine the validity of the conclusion. Your task is to assess whether the conclusion is "True," "False," or "Unknown" based on the information provided.

Task & Rules:
1. Premises: You will be provided with a set of statements or premises. These premises are the foundational truths or assumptions for the puzzle.
2. Conclusion: A statement will be presented as the conclusion. Your task is to evaluate this conclusion in the context of the given premises.
3. Evaluation Criteria:
   - True: The conclusion logically follows from the premises.
   - False: The conclusion contradicts the premises.
   - Unknown: The conclusion cannot be determined from the premises alone due to insufficient information.

Output Format:
- Please output your answer within a code block (```) as follows:
```
<result>
```
- Replace `<result>` with one of the following options: "True", "False", or "Unknown".

Puzzle:
premises:
- Elephantopus is a genus of perennial plants in the daisy family.
- Elephantopus is widespread over much of Africa, southern Asia, Australia, and the Americas
- Several species of Elephantopus are native to the southeastern United States.
- Elephantopus scaber is a traditional medicine.
conclusion: No Elephantopus is native to the southeastern United States.

Note: Ensure that your evaluation is based solely on the information provided in the premises without introducing external knowledge or assumptions.
\end{lstlisting}

\lstset{
    backgroundcolor=\color[RGB]{245,245,244},
    breaklines=true,
    xleftmargin=5pt, 
    xrightmargin=5pt, 
    breakindent=0pt,
    basicstyle=\ttfamily\scriptsize,
    frame=trbl,
    frameround = tttt,
    emph={Rules, Task, Output, Format, Puzzle},
    emphstyle={\bfseries\color{Goldenrod}}
}
\begin{lstlisting}[caption={Case of \texttt{Zebra Logic}},label=listing:case_zebra_logic]
You are tasked with solving a grid puzzle. This type of puzzle requires careful analysis of the provided background information and clues to deduce the correct arrangement of elements in a grid format. Follow the steps below to solve the puzzle and present your solution in the specified format.

Task & Rules:
1. Background Information: Carefully read any introductory information provided with the puzzle. This may include context or specific constraints that apply to the puzzle.
2. Clues: Analyze each clue given. These clues will guide you in determining the relationships between different elements in the grid.
3. Logical Deduction: Use logical reasoning to deduce the correct placement of each element in the grid. Consider all possible options and eliminate those that contradict the clues.
4. Consistency Check: Ensure that your solution is consistent with all the clues and background information provided.
Your response should include a solution followed by the final answer in a markdown table format. Use the following structure:
Assume column 1 is Year, column 2 is Wine, column 3 is Type.

Output Format:
- Please output your answer within a code block (```) as follows:
```
| 1984 | [Correct Wine] | [Correct Type] |
| 1988 | [Correct Wine] | [Correct Type] |
| 1992 | [Correct Wine] | [Correct Type] |
| 1996 | [Correct Wine] | [Correct Type] |
```

Puzzle:
Food: apricot, lemon  
Hobby: baking, card-games  
Job: bartender, writer  
Nationality: canadian, egyptian  

1. Nationality:canadian is on the left of Job:writer  
2. Hobby:card-games is on the right of Food:apricot  

Fill the following table to show your final answer.
| Food          | correct answer | correct answer |
| Hobby         | correct answer | correct answer |
| Job           | correct answer | correct answer |
| Nationality   | correct answer | correct answer |

You must stick to the given uncompleted table and must not transpose the table.
\end{lstlisting}

\end{document}